\documentclass[10pt,journal,compsoc]{IEEEtran}

\usepackage{enumitem}

\newlist{enumlist}{enumerate}{1}
\setlist[enumlist,1]{%
  label=(\arabic*), labelindent=0pt
}

\newlist{widelist}{enumerate}{1}
\setlist[widelist,1]{%
  label=(\arabic*), wide
}

\newlist{inlinelist}{enumerate*}{1}
\setlist*[inlinelist,1]{%
  label=(\roman*),
}

\newlist{Exlist}{enumerate}{1}
\setlist[Exlist,1]{%
  label=Ex\arabic*:, leftmargin=*, align=left
}

\usepackage[hidelinks]{hyperref}
\hypersetup{
    hypertex=true,
    colorlinks=false,
    linkcolor=blue,
    anchorcolor=blue,
    citecolor=blue
}

\usepackage{graphicx}

\usepackage{multirow}

\usepackage{stfloats}
\fnbelowfloat

\usepackage[capitalize]{cleveref}

\ifCLASSOPTIONcompsoc
  \usepackage[nocompress]{cite}
\else
  \usepackage{cite}
\fi
\ifCLASSINFOpdf
\else
\fi
\usepackage{array}
\begin{document}



%
\title{Facial Action Unit Detection and Intensity Estimation from Self-supervised Representation}




\author{
\IEEEauthorblockN{
    Bowen Ma\IEEEauthorrefmark{1},
    Rudong An\IEEEauthorrefmark{1},
    Wei Zhang,
    Yu Ding,
    \break
    Zeng Zhao,
    Rongsheng Zhang,
    Tangjie Lv,
    Changjie Fan,
    Zhipeng Hu
}

\IEEEcompsocitemizethanks{
    \IEEEcompsocthanksitem *Equal Contribution.
    \IEEEcompsocthanksitem Corresponding author: Yu Ding.
    \IEEEcompsocthanksitem B. Ma, R. An, W. Zhang, Y. Ding, Z. Zhao, R. Zhang, T. Lv, C. Fan, and Z. Hu are with the Netease Fuxi AI Lab, Hangzhou, China.
}
}


\IEEEtitleabstractindextext{%

\begin{abstract}

As a fine-grained and local expression behavior measurement, facial action unit (FAU) analysis  (e.g., detection and intensity estimation) has been documented for its time-consuming, labor-intensive, and error-prone annotation. Thus a long-standing challenge of FAU analysis arises from the data scarcity of manual annotations, limiting the generalization ability of trained models to a large extent. Amounts of previous works have made efforts to alleviate this issue via semi/weakly supervised methods and extra auxiliary information. However, these methods still require domain knowledge and have not yet avoided the high dependency on data annotation.
This paper introduces a robust facial representation model MAE-Face for AU analysis. Using masked autoencoding as the self-supervised pre-training approach, MAE-Face first learns a high-capacity model from a feasible collection of face images without additional data annotations. Then after being fine-tuned on AU datasets, MAE-Face exhibits convincing performance for both AU detection and AU intensity estimation, achieving a new state-of-the-art on nearly all the evaluation results. Further investigation shows that MAE-Face achieves decent performance even when fine-tuned on only 1\% of the AU training set, strongly proving its robustness and generalization performance.

\end{abstract}


}

\maketitle


\IEEEdisplaynontitleabstractindextext

\IEEEpeerreviewmaketitle


\ifCLASSOPTIONcompsoc
\IEEEraisesectionheading{
    \section{Introduction}
    \label{section:introduction}
}
\else
    \section{Introduction}
    \label{section:introduction}
\fi






%
%
%
%

\IEEEPARstart{F}{acial}
action units (FAU) are atomic descriptors introduced in the FACS\cite{FACS} for facial expression from the perspective of facial muscle movements. Nearly any facial expression can be represented by a certain combination of AUs with various AU intensities. For instance, a happy expression might be the combination of AU6 (Cheek Raiser) and AU12 (Lip Corner Puller), while a sad expression may involve AU4 (Brow Lowerer) and AU17 (Chin Raiser). Automatic facial action units analysis has drawn much attention from the affective computing community for its application potential in mental health\cite{lucey2010automatically}, human-computer interaction\cite{bevilacqua2016variations} and fatigue detection\cite{sikander2020novel}, etc.

The existing works have proposed to detect AU occurrence and estimate AU intensity in a face image.
Their basic idea is to extract facial features and then utilize them to classify the occurrence/absence or regress the intensity levels, based on supervised or semi-supervised learning. As known, supervised learning relies on high-quality but adequate training data. Unfortunately, available datasets consist of quite limited samples, since AU annotation is labor-intensive, time-consuming, and requires domain expertise. Due to the limitation of annotated AU samples, AU recognition may overfit on dataset labels and some other attributes like identity, background, illumination, etc. Thus some recent works\cite{yan2022weakly,niu2019multi,chang2022knowledge,zhang2021prior} exploit extra data via auxiliary tasks like expression recognition in a supervised manner to facilitate AU analysis. Though these works demonstrate their effectiveness and the significance of introducing extra in-the-wild data, they are still limited by the scale of extra-labeled data and the generalization ability of facial feature representation. Additionally, Chen \textit{et al.}\cite{chen2021understanding} have also documented that human annotation on facial expression is typically biased due to the personal, cultural, and societal differences, etc. between non-expert annotators. Thus when using these datasets with their labels we should take special care to prevent the model from learning these biases.


\begin{figure}[!t]
    \centering
    \includegraphics[width=9cm]{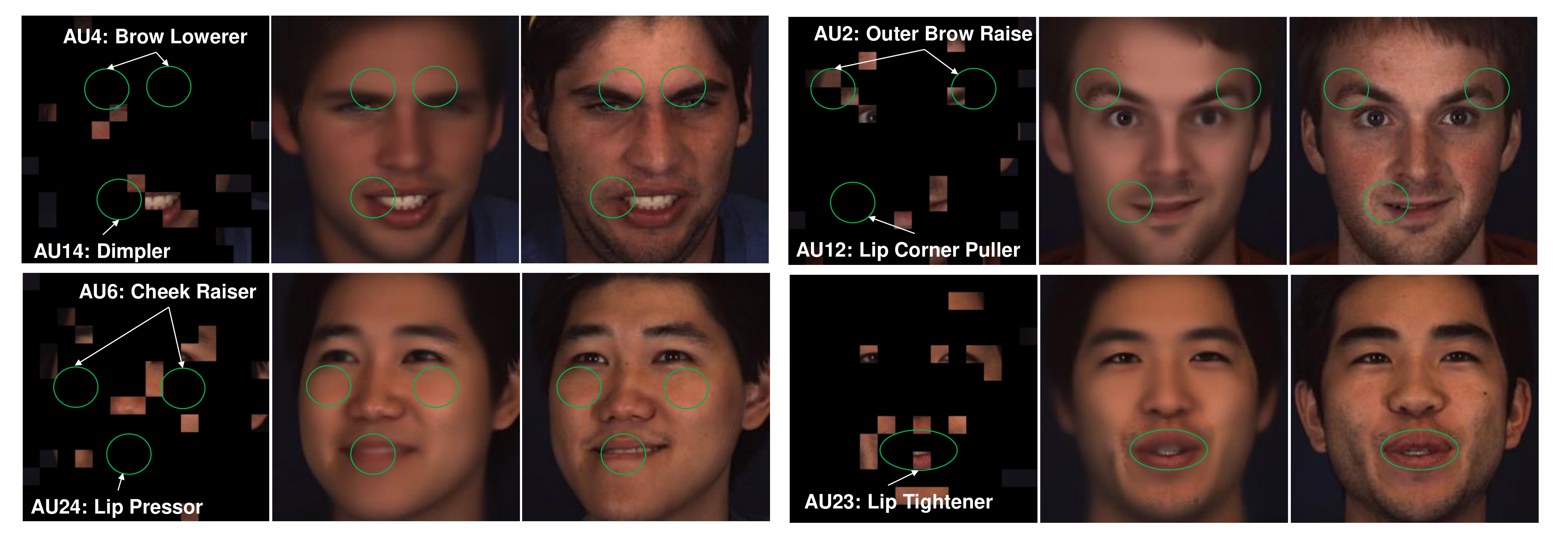}
    \caption{Reconstructed samples of our pre-trained MAE-Face on BP4D images. We show the 90\% masked image (left), our reconstruction (middle), and the original image (right). We observe that MAE-Face can reconstruct missing expression details corresponding to action units.}
    \label{fig:mae_reconstruction_bp4d}
\end{figure}
We seek to overcome these limitations via a self-supervised pre-training paradigm which dispenses with any annotation. Self-supervised learning has shown great success in both NLP\cite{BERT,GPT-3} and computer vision\cite{BEiT,MAE}. Due to no need for labeled data, self-supervised learning benefits from a large-scale dataset which can be easily acquired. The large-scale dataset allows for training a high-capacity model with rich representations and for making strong robustness and generalization performance in various downstream tasks.
Inspired by the success of these methods, we propose to pre-train a self-supervised model with a large set of face images. Unlike general vision models, our proposed model is intentionally designed for facial representation. It will be a great help in facilitating the generalization performance for AU-related tasks.

Motivated by the above analysis, we attempt to conduct a powerful pre-training method for facial representation.
Among the various techniques proposed for self-supervised visual representation learning, we opt to adopt masked autoencoder (MAE)\cite{MAE} as our baseline.
MAE proves that in image classification, it can outperform the other works when only using ImageNet-1k\cite{ImageNet} for training (which is not a very large-scale dataset nowadays).
Due to its high-ratio masking operation, MAE itself is served as strong data augmentation, boosting its generalization performance, which is beneficial for our work.
To this end, we build up our pre-training framework following the paradigm of MAE:
\begin{inlinelist}
\item we collect a bunch of facial datasets to pre-train the model dedicated to learning facial representations;
\item we fine-tune the pre-trained model on an AU dataset for AU detection or AU intensity estimation.
\end{inlinelist}
We name the pre-trained facial representation model \textbf{MAE-Face}. As shown in \cref{fig:mae_reconstruction_bp4d}, though most of the local regions corresponding to AUs are masked, our pre-trained MAE can still reconstruct it, revealing the great potential of facial representation learning.

The main contributions of our works can be summarized as follows:
\begin{enumlist}
\item We propose a robust facial representation model MAE-Face beneficial to automatic expression analysis, especially for facial action unit detection and intensity estimation. MAE-Face largely alleviates overfitting to the limited amount of AU-related data.
\item Our MAE-Face performs well and realizes new state-of-the-art results for AU detection and AU intensity estimation on three widely used datasets: BP4D, DISFA, and BP4D+. Even when fine-tuned with only 1\% of the training set, MAE-Face can still exhibit state-of-the-art results without much performance degradation.
\end{enumlist}

\section{Related Works}
\label{section:related_works}


Over the last few years, various works on AU detection and intensity estimation have been proposed. This section will represent the previous works about their contributions and deficiencies. Finally, we present some advanced self-supervised learning works highly related to our frameworks.

\subsection{AU Detection}

Conventional AU detection works often utilize hand-craft features like Gabor wavelet features\cite{valstar2006fully}, LBP\cite{jiang2011action,LP48}, SIFT\cite{zeng2015confidence} to detect AU by classifiers with Adaboost or SVM, etc. Recently deep learning methods boost the performance of AU detection to a new level. Considering the local region of AU, many previous works\cite{Onal2019D-PAttNet,JPML} detect AU based on patch learning either by dividing facial images uniformly\cite{DRML} or with the help of landmark detection\cite{JPML} and prior knowledge\cite{ma2019r}. More recently, Ge \textit{et al.}\cite{ge2021local} propose a muti-branch network to exploit the relationship among multiple patches and further introduce multi-level graph attention to enhance AU features learning\cite{ge2022mgrr}. These works make effort to efficiently extract facial features but neglect the data scarcity and thus are limited by the lack of data.

Due to the correlation that existed in AUs, the AU relationship modeling has been explored by lots of works\cite{tong2007facial,LP,corneanu2018deep,liu2020relation} to improve recognition accuracy.
Besides, with the success of attention mechanism, some recent works\cite{EAC-Net,Onal2019D-PAttNet,shao2021jaa,attnetionTrans2019,jacob2021facial} utilize it to enhance AU detection. These works demonstrate the effectiveness of AU relation modeling and attention learning, whereas they are merely trained on limited data which leads to poor generalization and restricts their performance.

To break through the limitation of data scarcity, some recent works consider introducing extra data or auxiliary information to aid AU detection. For instance, Cui \textit{et al.}\cite{cui2020knowledge} combine facial expression recognition with action unit detection, and train them jointly to refine their learning. Yang \textit{et al.}\cite{yang2021exploiting} using textual descriptions of AU occurrence to capture different AU semantic relations, exploiting semantic embedding and visual features for AU detection. Zhao \textit{et al.}\cite{zhao2018learning} utilize large-scale web images with inaccurate annotations to facilitate AU detection.  Zhang \textit{et al.}\cite{zhang2021prior,zhang2022transformer} introduce the pre-trained expression embedding \cite{zhang2021learning} feature as prior knowledge to enhance the AU detection.

More recently, Chen \textit{et al.}\cite{chen2022causal} focus on removing the effect caused by subject variation in AU recognition via a causal intervention module. To alleviate the reliance on densely annotated data, Li \textit{et al.}\cite{KS} propose to learn the interactive spatial-temporal correlation with extremely limited annotated data, consistency regularization, and pseudo-labeling to make the learning progress more efficient. However, these works still need extra data annotation or pseudo-labeling, thus the limitation caused by data scarcity has not yet been completely solved.

\subsection{AU Intensity}

Probabilistic models~\cite{rudovic2012multi,rudovic2012kernel,eleftheriadis2017gaussian} are extensively explored in AU intensity estimation. For instance, Walecki \textit{et al.}\cite{CCNN-IT} model AU co-occurrence patterns of AU intensity levels by combining conditional random field~(CRF).
Eleftheriadis \textit{et al.}\cite{VGP-AE} propose a Gaussian process auto-encoder for AU intensity estimation, which projects features via an encoder onto a latent space and reconstruct them to original features via a decoder. Markov Random Fields\cite{sandbach2013markov} and Conditional Ordinal Random Fields\cite{rudovic2012kernel,rudovic2012multi} are also been applied for predicting AU intensity.

Works\cite{KBSS,BORMIR,KJRE} leveraging deep models for AU intensity estimation also emerged in recent years.
Zhang \textit{et al.}\cite{KJRE} propose a novel weakly supervision training method with human knowledge to learn representation and intensity simultaneously. They also exploit another weakly-supervised method that learns frame-level intensity estimator and makes use of weakly labeled sequences data\cite{BORMIR}.
Sánchez-Lozano \textit{et al.}\cite{HR} jointly perform AU localization and intensity estimation through heatmap regression. Fan \textit{et al.}\cite{SCC-Heatmap} design a heatmap regression-based model and encode visual patterns of AUs in feature channels, reflecting the AU co-occurring pattern by the feature channels activation set.

More recently, context-aware temporal feature fusion and label fusion are applied in estimating AU intensity\cite{zhang2019context,APs}. Zhang \textit{et al.}\cite{zhang2019context} consider utilizing part of annotated databases and propose a weakly supervised patch-based framework for regressing AU intensity. They design a context-aware feature fusion module to capture spatial relationships among various local patches, and a context-aware label fusion module to capture the temporal dynamics. 
Sanchez \textit{et al.}\cite{APs} present a novel way to model temporal context based on neural processes, which contains global stochastic contextual representation, task-aware temporal context modeling, and selection.

\begin{figure*}[!t]
    \centering
    \includegraphics[width=18cm]{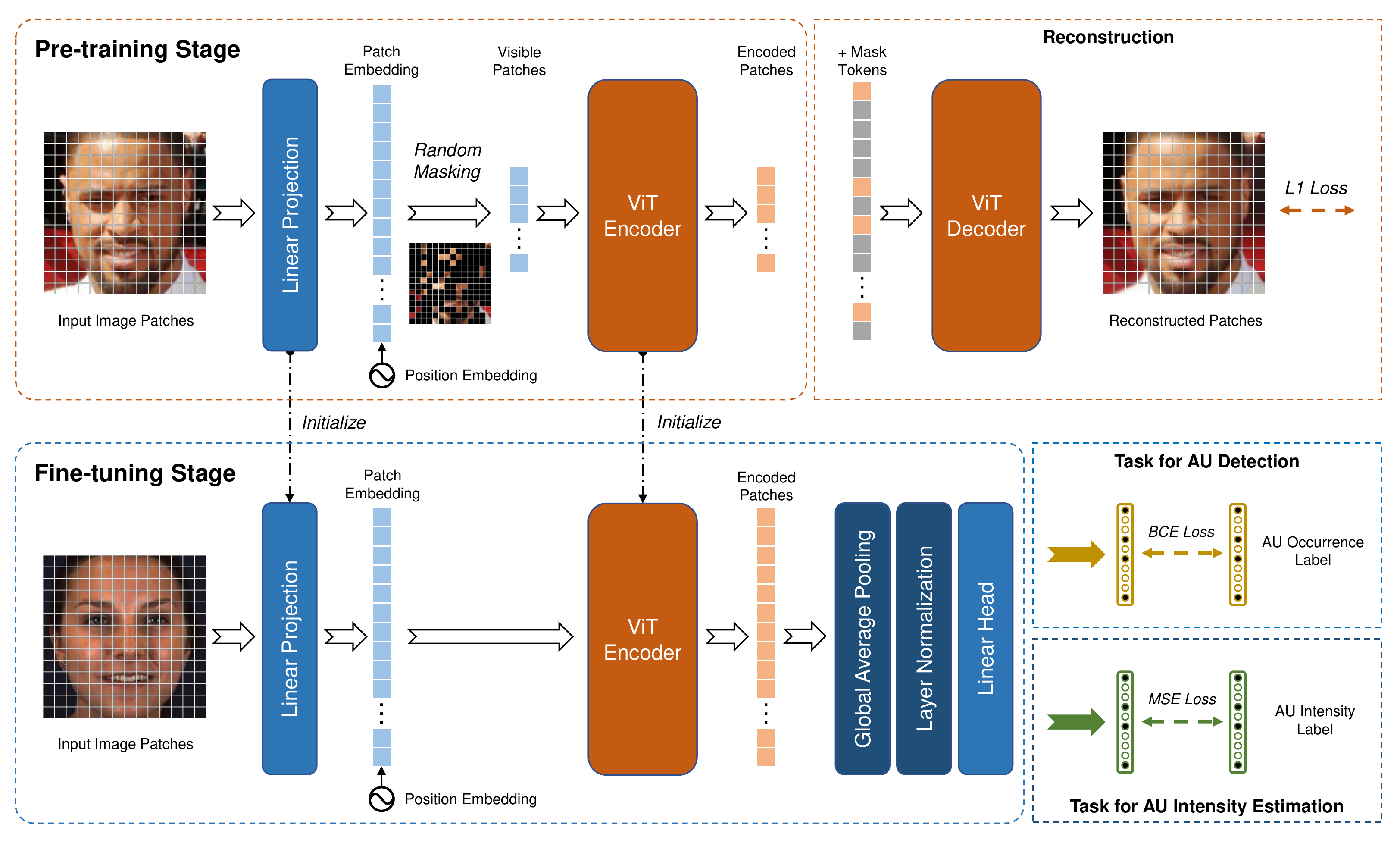}
    \vspace{-1em}
    \caption{The schematic pipeline of the proposed framework. It consists of two stages: the Pre-training Stage and the Fine-tuning Stage. The model learns facial representations from self-supervised pre-training for facial action unit detection and intensity estimation.}
    \label{fig:pipeline}
\end{figure*}

\subsection{Self-supervised Learning}

Recently, \textit{masked language modeling} has shown great success in self-supervised learning for NLP.
Auto-encoding models, such as BERT\cite{BERT}, are pre-trained to reconstruct missing contents corrupted by random masking.
Auto-regressive models, such as GPT-3\cite{GPT-3}, are pre-trained to predict the next token based on all the previous ones.
These models benefit from self-supervised learning to greatly improve their performance by utilizing very large-scale datasets.

Meanwhile, \textit{contrastive learning}\cite{wu2018unsupervised,oord2018representation,hjelm2018learning,bachman2019learning,he2020momentum,chen2020simple} manages to learn a visual representation in a self-supervised manner.
It proposes to learn from the images using carefully designed augmentations, where the model tries to represent the positive pairs (same image with different augmentations) as close as possible and the negative pairs (different images) as far as possible.
In this way, it learns to extract high-level semantic features not influenced by the augmentations.

\textit{Masked image modeling}, similar to masked language modeling, has been tried but not showing many successes\cite{ViT}.
Until recently, following the NLP routines using tokens, BEiT\cite{BEiT} comes up with a pre-training framework to predict visual tokens of the missing patches with masked image patches as the input.
As an alternative, MAE\cite{MAE} proposes to directly reconstruct pixels of the masked patches, which is a more straightforward method natively designed for image modeling.

BEiT uses a pre-trained discrete VAE (dVAE) from DALL-E\cite{DALLE} to generate predicted tokens, where the dVAE is pre-trained with extra data (250M images).
Since we want to build a visual representation for face images specifically, the dVAE trained with general images\cite{DALLE} may not be the optimal tokenizer for our problem.
Besides, in BEiT pre-training, the dVAE inference also brings extra computation.
With no need for a tokenizer, MAE is much simpler to implement and also faster to train.
Thus, we build our pre-training framework based on MAE as the self-supervised visual representation learner.


\section{Method}
\label{section:method}

In this section, we will introduce our proposed method in detail. Firstly, we describe the problem definition of AU detection and intensity estimation and then represent the overfitting problems that commonly exist in previous works. Then, we give an overview of our proposed framework. After that, we introduce the pre-training and fine-tuning procedure. Finally, we present the training loss used in our work.

\begin{figure*}[!t]
    \centering
    \includegraphics[width=18cm]{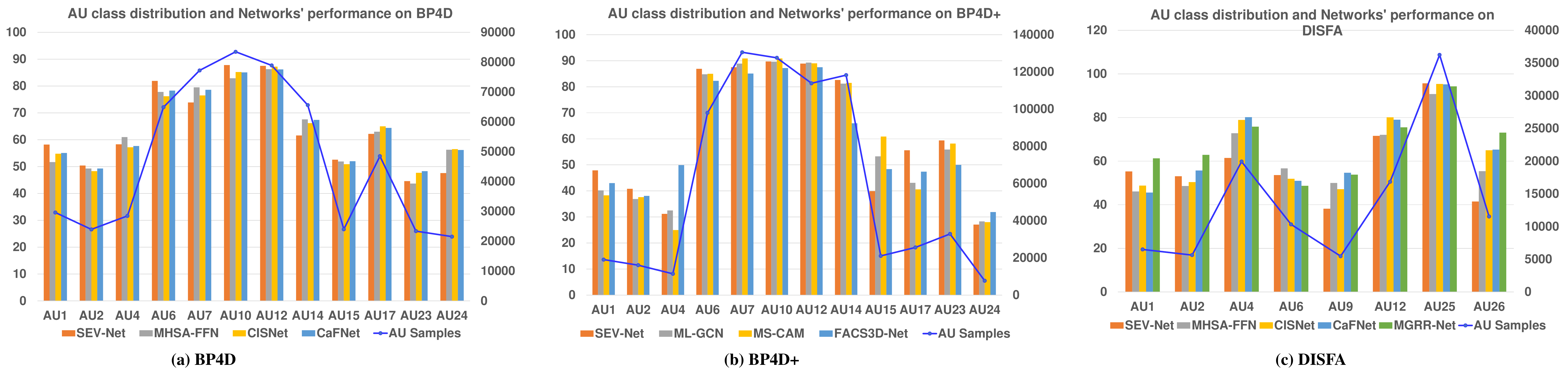}
    \caption{The F1 score on AU detection for several state-of-the-art works (SEV-Net\cite{yang2021exploiting}, MHSA-FFN\cite{jacob2021facial}, CIS-Net\cite{chen2022causal}, CaFNet\cite{chen2021cafgraph} and MGRR-Net\cite{ge2022mgrr}) and the AU class sample distributions (blue lines), for BP4D, BP4D+ and DISFA, respectively, from which we can see that the F1 score on each AU of all methods are consistent with the AU class distributions.}
    \label{fig:dataset_network_performance}
\end{figure*}

\subsection{Problem Definition}

\begin{figure}[!t]
    \centering
    \includegraphics[width=8.5cm]{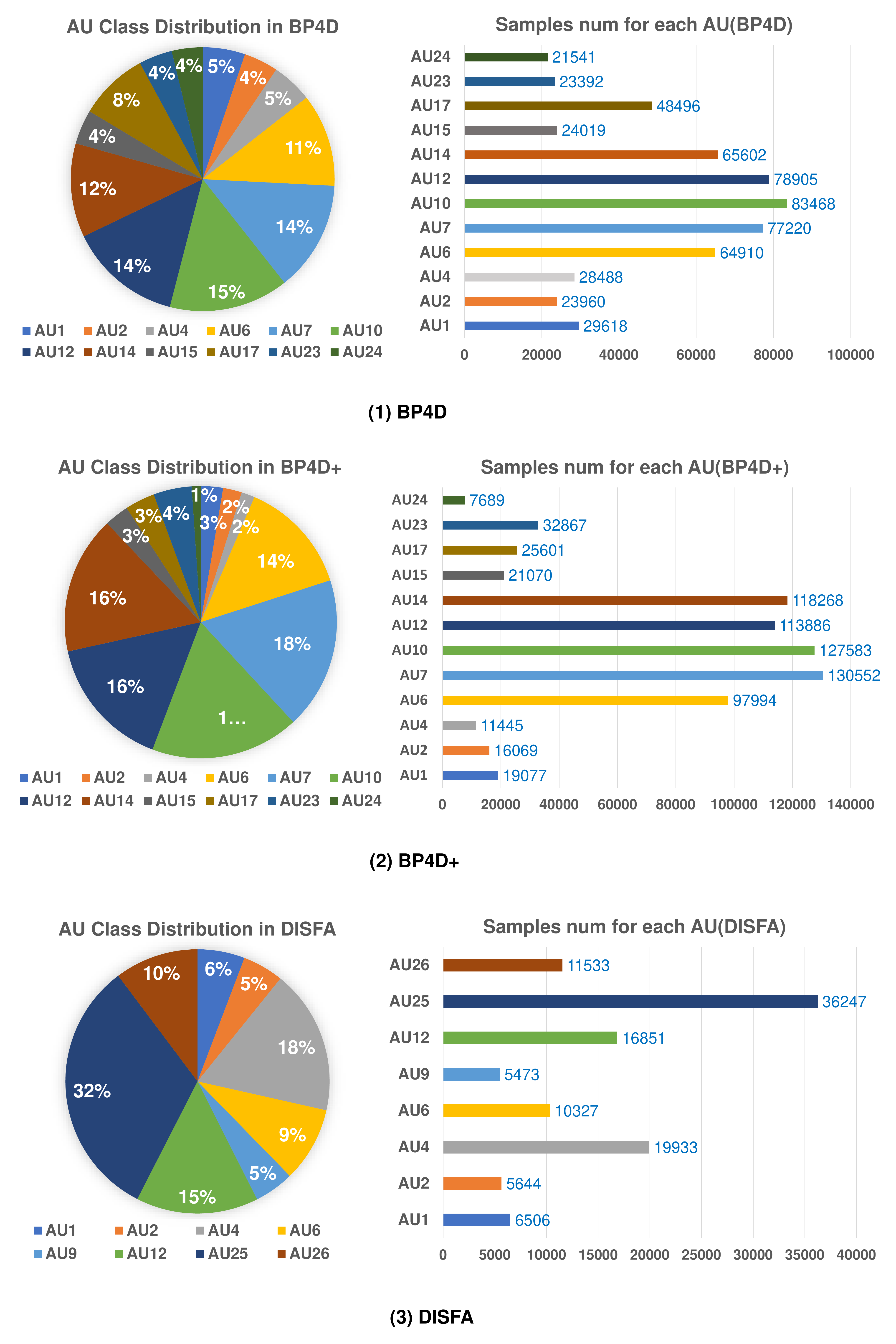}
    \caption{The AU classes distributions for BP4D, BP4D+ and DISFA.}
    \label{fig:dataset_stastic_new_3}
\end{figure}
As mentioned above, this work addresses both problems of AU detection and intensity estimation. Specifically, AU detection aims to recognize the occurrence or absence of AUs in a face image. Since there are multiple AUs in an input image, AU detection is a multi-label classification problem. It aims to perform the AU binary classification (i.e. 0 for absence and 1 for presence) for n AUs: $P=[p_1,...,p_i,...,p_n]$, where $p_i \in {0,1}$ and $n$ is the total number of AUs in the target dataset. Moreover, the AU intensity estimation is defined as six discrete levels from 0 to 5 that indicates various degrees of facial action movements, where 0 and 5 indicate the lowest and the highest activation degree, respectively. The target of AU intensity estimation task is to regress the AU intensity levels for n AUs, which is termed as $L=[l_1,...,l_i,...,l_n]$, where $l_i \in {0,1,...,5}$.

We've mentioned in Section~\ref{section:introduction} that AU detection or intensity estimation suffers from overfitting to the training set due to the limited amount of available data in the released datasets. Furthermore, our work attempts to recognize the mentioned overfitting as two aspects, including the overfitting to AU labels and the overfitting to identities, which will be detailed as follows.

\textit{Overfitting to AU labels.}
It is well-known that data imbalance is a common issue for image recognition\cite{chou2020remix,liu2019large,cui2019class}, which may bias neural networks towards the majority classes when the training data is heavily imbalanced. Similarly, AU-related tasks also suffer from this problem due to the imbalance of label distribution in the available datasets. \cref{fig:dataset_stastic_new_3} show the distributions of AU classes from three commonly-used datasets (please refer to Section~\ref{section:experiments} for more details). We can observe the imbalance of the distributions of AU classes. For example, \cref{fig:dataset_stastic_new_3} (1) and (2) show that samples of AU6,7,10,12,14 are observably far more than the other AUs in BP4D and BP4D+; \cref{fig:dataset_stastic_new_3} (3) shows that AU4,12 and AU25 are the dominant classes in DISFA. On the other hand, \cref{fig:dataset_network_performance} shows the AU detection performance of previous works (please refer to \cref{table:results:au_det_bp4d,table:results:au_det_bp4d+,table:results:au_det_disfa} for the results from more existing works.). It can be observed that the performance (please refer to \cref{table:results:au_det_bp4d,table:results:au_det_bp4d+,table:results:au_det_disfa}) is consistent with the AU class distributions in \cref{fig:dataset_stastic_new_3}. In the BP4D dataset, the performance of previous works on the majority classes (AU6, 7, 10, 12) tends to be much higher than that on the minority classes (AU1, 2, 4, 15, 23, 24). It shows that the previous works suffer from the problem of overfitting the dominant AU classes.

\textit{Overfitting to identities.}
It is expected to move out identity information from the expression latent representations for the AU tasks\cite{chen2022causal,xiang2017linear}. Unfortunately, some expression are scarce and only appear on a few specific identities, leading to the entanglement of these expressions and identities. Moreover, this entanglement is also aggravated, as the identities of available AU datasets (41, 140 and 27 subjects in BP4D, BP4D+ and DISFA respectively) are too few. Similar to identities, some other attributes unrelated to expression, like background, illumination, view, pose and makeup, are hard to leave out from the expression latent representations for AU tasks. Similar issue also appears in facial expression recognition\cite{li2020deep}, since their scarcity leads to their entanglement with specific expression.

\subsection{Framework}

Due to the above-mentioned limitation regarding the data, training directly on AU datasets from scratch will inevitably fall into the dilemma of overfitting.
To address the overfitting issue, we seek to build a good facial representation model that can better generalize to facial expression.

Therefore, instead of one-stage training, we carry out a two-stage framework.
Our model is pre-trained on facial datasets in Stage~1, and then fine-tuned on AU datasets in Stage~2. We name the pre-trained model \textbf{MAE-Face}.

Compared to AU datasets, large-scale in-the-wild facial datasets exhibits much more diversity in terms of identities, expressions, makeups, and photographic attributes.
By utilizing the two-stage framework, MAE-Face benefits from in-the-wild datasets to learn a robust facial representation in Stage~1.
Then in Stage~2, MAE-Face is further optimized to learn a task-specific predictor for action units.

This section will introduce the proposed framework in detail, including the pre-training, the fine-tuning, and the data pre-processing used.

\subsubsection{Facial representation pre-training}
\label{section:method:mae_pretrain}




As mentioned above, MAE-Face benefits from large-scale datasets to learn a robust facial representation.
To avoid any human-labor annotations for the data, we opt to adopt a self-supervised learning method for the pre-training, with which we can easily utilize a large set of in-the-wild face images.

\begin{figure}[!t]
    \centering
    \includegraphics[width=9cm]{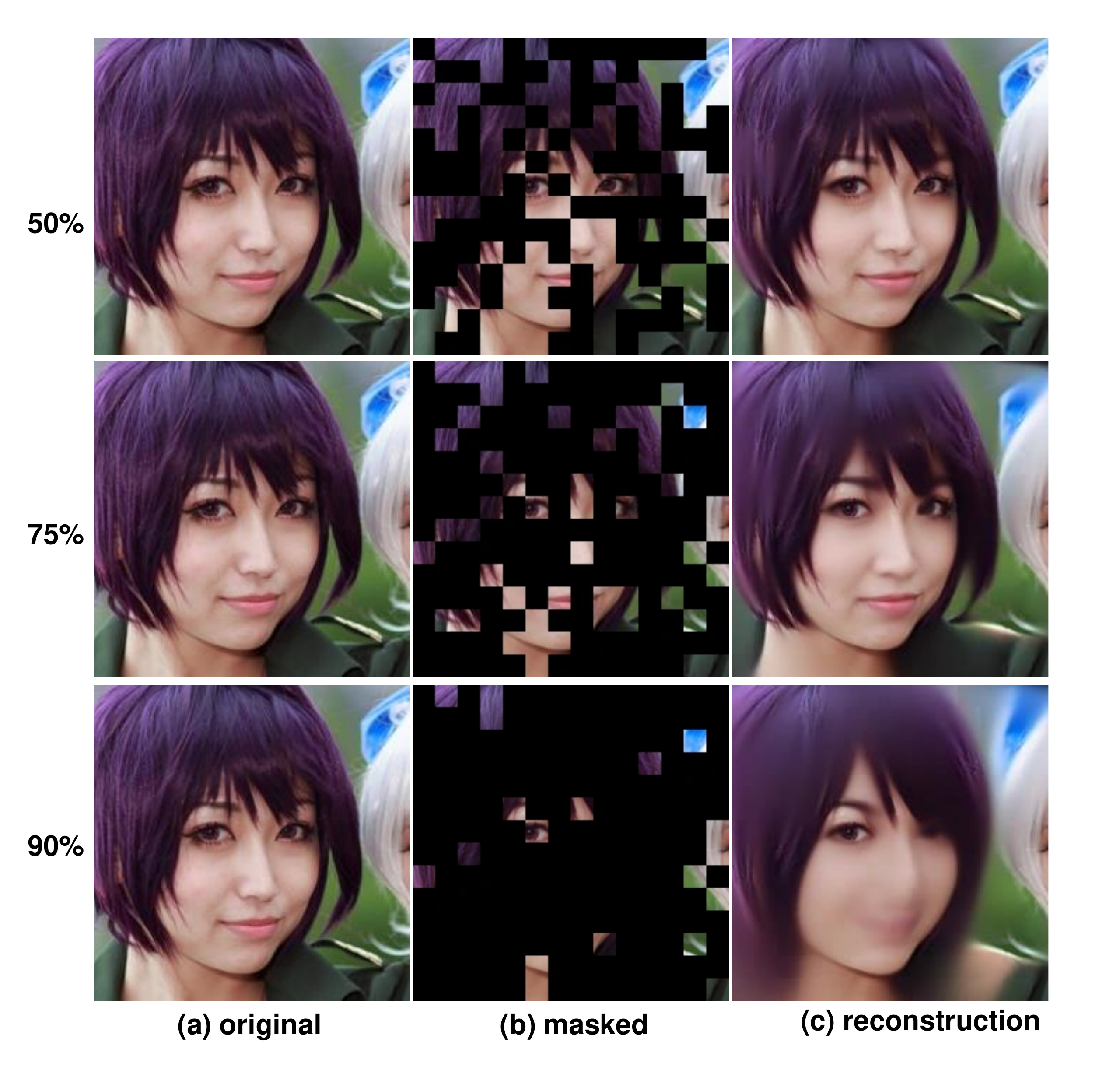}
    \vspace{-2em}
    \caption{Samples of different masking ratios. When masking out 90\% patches, MAE-Face fails to reconstruct fine-grained facial details.}
    \label{fig:mask_dataset}
\end{figure}

Among the various self-supervised vision learning methods, masked autoencoding~\cite{ViT,BEiT,MAE} is used as our pre-training paradigm.
In masked autoencoding, the model learns to predict the missing contents given a partial observation of an image.
Fig.~\ref{fig:mask_dataset} shows some examples of the reconstructed results by MAE-Face.
From these examples, we conclude that to obtain a global view from a local patch, the model has learned a high-level understanding and reasoning ability for facial image contents.
It implies that MAE-Face has a strong facial representational capability to facilitate the performance of downstream tasks.

\textbf{Architecture} of our pre-training procedure is illustrated in Fig.~\ref{fig:pipeline} (\textit{Pre-training Stage}), where the masked autoencoder is constructed based on Vision Transformer (ViT)\cite{ViT}.
It includes a ViT-Base encoder and a ViT decoder, where the encoder has a depth of 12 and a width of 768, and the decoder has a depth of 8 and a width of 512.

\textbf{Masking.}
Specifically, the pre-training follows a masking-then-reconstruct procedure\cite{MAE}.
Before feeding into the model, an image is uniformly split into non-overlapping patches of $16\times16$.
Following \cite{Transformer}, patch embedding is constructed by a linear projection.
Position embedding is added to mark the relative position of the patches.
Then, a portion of the patches is masked out and the rest is used as the input, and the training target is to reconstruct the missing patches in pixels from this partial observation.
As shown in Fig.~\ref{fig:mask_dataset}, MAE-Face can reconstruct fine-grained details well from the masked face image of 50\% or 75\%, but fails in 90\%.
Therefore for the pre-training, we choose to use a mask ratio of 75\% as a balanced choice.

\textbf{Initialization.}
Instead of random initialization, we initialize the ViT encoder using the ImageNet-1k pre-trained encoder released by \cite{MAE}. It will speed up the pre-training procedure, where a 200-epoch pre-trained model can exhibit decent performance in downstream tasks (e.g. in Section~\ref{section:ablation:pretrain_loss_function}). Nevertheless, we can still benefit from a longer pre-training, so we use 800 epochs in most of our experiments.

\textbf{Dataset.}
The pre-training dataset is built from 4 datasets:
AffectNet\cite{AffectNet}, CASIA-WebFace\cite{CASIA-WebFace}, IMDB-WIKI\cite{IMDB-WIKI} and CelebA\cite{CelebA}.
These datasets are concatenated to form a large set of 2,170,000 face images, with all the labels removed.
Nowadays, compared with the datasets used in numerous large-scale pre-training, the proposed dataset is relatively small.
We suppose that the random masking also serves a strong data augmentation, regularizing the training procedure and making masked autoencoding a data-efficient method.
Thus, benefiting from masked autoencoding, MAE-Face can still learn a good facial representation on this relatively small dataset.

\textbf{Pre-processing for the pre-training dataset.}
\begin{inlinelist}
\item The dataset is cleaned up by dropping corrupt files and images with too low resolution;
\item RetinaFace\cite{RetinaFace} is applied to detect 5-point facial landmarks, where the coordinates of the two eyes are used for face alignment;
\item RetinaFace is applied again to detect the facial bounding box and crop a squared image.
\end{inlinelist}

\subsubsection{End-to-end fine-tuning on Action Units}

In Stage~2, the pre-trained model is fine-tuned on AU datasets for AU detection or AU intensity estimation.
Following \cite{MAE}, we apply an end-to-end fine-tuning mechanism instead of linear probing to obtain the highest performance.

We drop the pre-trained ViT \textit{decoder} because it is designed for reconstructing pixels but not for downstream tasks.
However, after learning a robust facial representation, the pre-trained ViT \textit{encoder} serves as the key component in our framework.



\textbf{Classifier.}
As shown in \cref{fig:pipeline} (\textit{Stage2: AU-Related Task}), we build up our classifier or estimator using the pre-trained ViT encoder, followed by a global average pooling, a Layer Normalization\cite{LayerNorm} and a linear projection head for the classification or estimation.
The encoder is supposed to learn a facial expression representation and the head is supposed to learn a linear mapping to the final output. Specifically, the model is fine-tuned for each downstream task independently.

\textbf{Fine-tuning on AU datasets.}
During fine-tuning, all image patches are fed into the model without masking, and the optimization target is to detect AU occurrence or estimate AU intensity. Fine-tuning for AU detection is conducted on BP4D, BP4D+, and DISFA, while for AU intensity, following previous works, only on BP4D and DISFA (please ref to Section~\ref{section:experiments} for more details).
The encoder and the linear head are optimized together using backpropagation.
Finally, the optimization iteration stops when the validation loss stops decreasing.

\textbf{Pre-processing for the AU datasets.}
Same as in pre-training, face alignment and face cropping are applied to extract aligned face images from the original AU datasets.
In common practices, face alignment and face cropping plays an important role in improving the performance of facial recognition tasks.
We follow this practice both in pre-training and fine-tuning to further boost the performance of our proposed method.

\subsection{Loss Functions}

Our pre-training procedure is a masking-then-reconstruct task, which aims to predict the masked patches. The loss of the pre-training task is $\mathcal{L}_{pre}$. We then validate our pre-trained model on both AU detection and AU intensity estimation tasks, whose losses are denoted as $\mathcal{L}_{det}$ and $\mathcal{L}_{int}$, respectively, each of them is introduced in detail in the following.

\textbf{Loss for pre-training.}
Instead of the L2 loss proposed by \cite{MAE}, the model pre-trained using L1 loss exhibits better performance both in reconstruction and downstream tasks (see Section~\ref{section:ablation:pretrain_loss_function}).
Suppose there are $K$ masked patches in the input. To predict the $K$ masked patches, we utilize pixel-wise L1 loss to constrain the distance between the prediction ($\hat m_k$) and ground-truth ($m_k$) of the masked patches. Thus $\mathcal{L}_{pre}$ can be denoted as:
\begin{equation}
\mathcal L_{pre} = \sum_{k=1}^{K} || m_k - \hat m_k ||,
\end{equation}

\textbf{Loss for AU detection.}
To constrain the AU detection results to be as close as possible to the AU ground-truth, we formulate it as a multi-label binary classification problem, where each AU label is a binary. Thus $\mathcal{L}_{det}$ is defined as:
\begin{equation}
\mathcal L_{det} = -\sum_{i=1}^{N} [p_i log \hat{p}_i + (1 - p_i) log(1-\hat{p}_i)],
\end{equation}
where $N$ is the number of AUs, $\hat p_i$ denotes the predicted probability which is the model output activated by a sigmoid function, and $p_i$ denotes the binary ground-truth of the $AU_{i}$.

\textbf{Loss for AU intensity estimation.}
Since the AU intensity values are annotated as discrete integers in 0$\sim$5, we normalize the ground-truth values to 0$\sim$1. Then, the goal of our model is to regress the normalized AU intensity values. We use L2 distance to minimize the distance between the predicted value $\hat v_i$ and ground-truth $v_i$ for $AU_{i}$, thus $\mathcal{L}_{int}$ is denoted as:
\begin{equation}
\mathcal L_{int} = \sum_{i=1}^{N} || v_i - \hat v_i ||^2_2,
\end{equation}
During inference, the model output is scaled to the original 0$\sim$5 range for evaluation.

\section{Experiments}
\label{section:experiments}


This section reports extensive experiments to validate the effectiveness of our method and discuss the experimental results.
These experiments are carried out on three public benchmarks commonly used for AU detection and intensity estimation.
We first present the datasets in detail and then describe implementation settings, evaluation metrics, comparisons with state-of-the-art methods, and ablation studies.
Also, we provide some visualized results to demonstrate the advantages of our approach.
To further obtain insights into our method, we design ablation experiments to study the impact of different pre-trained models and loss functions.



\textbf{Datasets.}
We employ the commonly used AU datasets BP4D\cite{BP4D}, BP4D+\cite{BP4D+}, and DISFA\cite{DISFA} to perform AU detection and AU intensity estimation.
BP4D contains about 147,000 frames from 328 video clips. These videos record the spontaneous expressions of 41 subjects~(23 female and 18 male) when they perform 8 tasks. Each frame is annotated with the occurrence or absence of 12 AUs for AU detection and the intensity of 5 AUs for AU intensity estimation.
BP4D+ contains about 197,000 frames from 1400 video clips. These videos are from 140 subjects~(82 females and 58 males). Like BP4D, each frame of BP4D+ is annotated with the occurrence or absence of 12 AUs for AU detection.
DISFA contains about 131,000 frames from 27 video clips. These videos are from 27 subjects~(12 females and 15 males). Each frame of DISFA is annotated with the occurrence or absence of 8 AUs for AU detection and the intensity of 12 AUs for AU intensity estimation.

We perform AU detection on BP4D, BP4D+, and DISFA and AU intensity estimation on BP4D and DISFA, which follow the previous works for experimental comparisons.

\textbf{Evaluation metrics.}
\textit{For AU detection}, we evaluate the model performance with F1 score, which is defined as the harmonic mean of precision and recall:
\begin{equation}
F1 = 2 \times \frac{Precision \cdot Recall}{Precision+Recall}
\end{equation}
The F1 score is designed for evaluating the model classification performance on the imbalanced dataset by taking the precision-recall trade-off into account.

\textit{For AU intensity estimation}, we evaluate the model performance with intra-class correlation (ICC)\cite{ICC}, mean squared error (MSE) and mean absolute error (MAE).
ICC measures the agreement of ratings by comparing the variability of different ratings of the same subject to the total variation across all ratings and all subjects.
Following the previous works, we use the ICC (3,1) variant.
Mean squared error and mean absolute error are the common metrics used to measure the average difference between the prediction and the target in regression problems.

\textit{For all metrics above}, we evaluate and report them on each AU. We also calculate the average metric value of all AU categories to achieve a comprehensive result.
For fair comparisons in AU detection, our evaluation follows the prior works\cite{li2018eac,shao2021jaa}, using 3-fold cross-validation and also their specific splits of the training set and the validation set.
To evaluate AU intensity estimation on BP4D, following\cite{walecki2017deep,SCC-Heatmap,attnetionTrans2019} on the FERA2015 training and development partition, we train our method on 21 subjects and test on 20 subjects. To evaluate AU intensity estimation on DISFA, we also perform subject-exclusive 3-fold cross-validation like the previous works\cite{walecki2017deep,SCC-Heatmap,attnetionTrans2019}.

\begin{table*}[!t]

\setlength{\tabcolsep}{6pt}
\renewcommand{\arraystretch}{1.3}
\renewcommand{\extrarowheight}{0.5pt}

\caption{Comparison of AU detection results on BP4D in terms of F1 scores (\%)}
\label{table:results:au_det_bp4d}
\centering

\begin{tabular}{l|cccccccccccc|c}
    \hline
    Method & AU1 & AU2 & AU4 & AU6 & AU7 & AU10 & AU12 & AU14 & AU15 & AU17 & AU23 & AU24 & Avg. \\[1pt]
    \hline
    \hline
    LP-Net\cite{LP}  &43.4 &38.0 &54.2 &77.1 &76.7 &83.8 &87.2 &63.3 &45.3 &60.5 &48.1 &54.2 &61.0 \\
    ARL\cite{attnetionTrans2019}     &45.8 &39.8 &55.1 &75.7 &77.2 &82.3 &86.6 &58.8 &47.6 &62.1 &47.4 &55.4 &61.1 \\
    J{\^A}A-Net\cite{shao2021jaa} &53.8 &47.8 &58.2 &78.5 &75.8 &82.7 &88.2 &63.7 &43.3 &61.8 &45.6 &49.9 &62.4 \\
    SRERL\cite{SRERL}   &46.9 &45.3 &55.6 &77.1 &78.4 &83.5 &87.6 &63.9 &52.2 &63.9 &47.1 &53.3 &62.9 \\
    AU-RCNN\cite{ma2019r} &50.2 &43.7 &57.0 &78.5 &78.5 &82.6 &87.0 &67.7 &49.1 &62.4 &[50.4] &49.3 &63.0 \\
    UGN-B\cite{song2021uncertain}   &54.2 &46.4 &56.8 &76.2 &76.7 &82.4 &86.1 &64.7 &51.2 &63.1 &48.5 &53.6 &63.3 \\
    HMP-PS\cite{song2021hybrid} &53.1 &46.1 &56.0 &76.5 &76.9 &82.1 &86.4 &64.8 &51.5 &63.0 &49.9 &54.5 &63.4 \\
    LGRNet\cite{ge2021local} &50.8 &47.1 &57.8 &77.6 &77.4 &84.9 &88.2 &66.4 &49.8 &61.5 &46.8 &52.3 &63.4 \\
    MGRR-Net\cite{ge2022mgrr} &52.6 &47.9 &57.3 &78.5 &77.6 &84.9 &88.4 &[67.8] &47.6 &63.3 &47.4 &51.3 &63.7 \\
    SEV-Net\cite{yang2021exploiting} &[58.2] &[50.4] &58.3 &\textbf{81.9} &73.9 &\textbf{87.8} &87.5 &61.6 &52.6 &62.2 &44.6 &47.6 &63.9 \\
    MHSA-FFN\cite{jacob2021facial} &51.7 &49.3 &[61.0] &77.8 &79.5 &82.9 &86.3 & 67.6 &51.9 &63.0 &43.7 &[56.3] &64.2 \\
    CISNet\cite{chen2022causal} &54.8 &48.3 &57.2 &76.2 &76.5 &85.2 &87.2 &66.2 &50.9 &[65.0] &47.7 &\textbf{56.5} &64.3 \\
    D-PAttNe$t^{tt}$\cite{Onal2019D-PAttNet} &50.7 &42.5 &59.0 &79.4 &79.0 &85.0 &\textbf{89.3} &67.6 &51.6 &\textbf{65.3} &49.6 &54.5 &64.7 \\
    CaFNet\cite{chen2021cafgraph} &55.1 &49.3 &57.7 &78.3 &78.6 &85.1 &86.2 &67.4 &52.0 &64.4 &48.3 &56.2 &64.9 \\
    ME-GraphAU\cite{ME-GraphAU} &52.7 &44.3 &60.9 &[79.9] &\textbf{80.1} &85.3 &[89.2] &\textbf{69.4} &\textbf{55.4} &64.4 &49.8 &55.1 &[65.5] \\
    \hline
    \textbf{MAE-Face} & \textbf{62.5} & \textbf{56.4} & \textbf{66.3} & 79.6 & [79.6] & [85.6] & 89.1 & 64.2 & [54.5] & [65.0] & \textbf{53.8} & 51.8 & \textbf{67.4} \\
    \hline
\end{tabular}

\centering
~\\
The best results are shown in bold, and the second best in brackets.

\end{table*}

\begin{table*}[!t]

\setlength{\tabcolsep}{6pt}
\renewcommand{\arraystretch}{1.3}
\renewcommand{\extrarowheight}{0.5pt}

\caption{Comparison of AU detection results on BP4D+ in terms of F1 scores (\%)}
\label{table:results:au_det_bp4d+}
\centering

\begin{tabular}{l|cccccccccccc|c}
    \hline
    Method & AU1 & AU2 & AU4 & AU6 & AU7 & AU10 & AU12 & AU14 & AU15 & AU17 & AU23 & AU24 & Avg. \\[1pt]
    \hline
    \hline
    FACS3D-Net\cite{yang2019facs3d} &43.0 &38.1 &\textbf{49.9} &82.3 &85.1 &87.2 &87.5 &66.0 &48.4 &47.4 &50.0 &[31.9] &59.7 \\
    ML-GCN\cite{yang2019facs3d} &40.2 &36.9 &32.5 &84.8 &[88.9] &89.6 &[89.3] &81.2 &[53.3] &43.1 &55.9 &28.3 &60.3\\
    MS-CAM\cite{yang2019facs3d} &38.3 &37.6 &25.0 &85.0 &\textbf{90.9} &\textbf{90.9} &89.0 &[81.5] &\textbf{60.9} &40.6 &58.2 &28.0 &60.5 \\
    SEV-Net\cite{yang2021exploiting} &[47.9] &[40.8] &31.2 &\textbf{86.9} &87.5 &89.7 &88.9 &\textbf{82.6} &39.9 &\textbf{55.6} &[59.4] &27.1 &[61.5] \\
    \hline
    \textbf{MAE-Face} & \textbf{52.5} & \textbf{42.9} & [44.1] & [86.1] & [88.9] & [90.8] & \textbf{90.2} & 81.1 & 48.0 & [50.6] & \textbf{61.2} & \textbf{41.1} & \textbf{64.8} \\
    \hline
\end{tabular} 

\centering
~\\
The best results are shown in bold, and the second best in brackets.

\end{table*}

\begin{table}[!t]

\vspace{1em}
\setlength{\tabcolsep}{0.5pt}
\renewcommand{\arraystretch}{1.5}
\renewcommand{\extrarowheight}{0.5pt}

\caption{Comparison of AU detection results on DISFA in terms of F1 scores (\%)}
\label{table:results:au_det_disfa}
\centering

\begin{tabular}{l|cccccccc|c}
    \hline
    Method & AU1 & AU2 & AU4 & AU6 & AU9 & AU12 & AU25 & AU26 & Avg. \\[1pt]
    \hline
    \hline
    AU-RCNN    &32.1 &25.9 &59.8 &55.4 &39.8 &67.7 &77.4 &52.6 &51.3 \\
    SRERL      &45.7 &47.8 &56.9 &47.1 &45.6 &73.5 &84.3 &43.6 &55.9 \\
    SEV-Net    &55.3 &53.1 &61.5 &53.6 &38.2 &71.6 &[95.7] &41.5 &58.8 \\
    UGN-B      &43.3 &48.1 &63.4 &49.5 &48.2 &72.9 &90.8 &59.0 &60.0 \\
    HMP-PS     &38.0 &45.9 &65.2 &50.9 &50.8 &76.0 &93.3 &67.6 &61.0 \\
    MHSA-FFN   &46.1 &48.6 &72.8 &[56.7] &50.0 &72.1 &90.8 &55.4 &61.5 \\
    ME-GraphAU &54.6 &47.1 &72.9 &54.0 &[55.7] &76.7 &91.1 &53.0 &63.1 \\
    J{\^A}A-Net    &62.4 &60.7 &67.1 &41.1 &45.1 &73.5 &90.9 &67.4 &63.5 \\
    CISNet     &48.8 &50.4 &[78.9] &51.9 &47.1 &\textbf{80.1} &95.4 &65.0 &64.7 \\
    CaFNet     &45.6 &55.7 &\textbf{80.2} &51.0 &54.7 &[79.0] &95.2 &65.3 &65.8 \\
    LGRNet     &[62.6] &\textbf{64.4} &72.5 &46.6 &48.8 &75.7 &94.4 &[73.0] &67.3 \\
    MGRR-Net   &61.3 &[62.9] &75.8 &48.7 &53.8 &75.5 &94.3 &\textbf{73.1} &[68.2] \\
    \hline
    \textbf{MAE-Face} & \textbf{68.4} & 59.4 & 76.5 & \textbf{58.4} & \textbf{56.7} & 78.5 & \textbf{96.6} & 71.7 &\textbf{70.8} \\
    \hline
\end{tabular}

\centering
~\\
The best results are shown in bold, and the second best in brackets.

\end{table}


\textbf{Experimental settings.}
\textit{For the pre-training settings}, we pre-train MAE-Face for 800 epochs, with 40 warmup epochs. AdamW optimizer\cite{AdamW} is used with a weight decay of 0.05. Random cropping is applied as the data augmentation. The Transformer blocks are initialized with Xavier Uniform\cite{Xavier}.
We set the base learning rate $LR_{base}$ to 1.5e-4. The actual learning rate is derived using the linear \textit{LR} scaling rule\cite{large_batch}:
\begin{equation}
LR = LR_{base} \times \frac{BatchSize}{256}
\end{equation}
where we set the batch size to 4096. The learning rate is scheduled with cosine decay\cite{SGDR}.
The framework is constructed using PyTorch\cite{PyTorch} and trained on 8 NVIDIA A30 GPUs.

\textit{For the fine-tuning settings}, we fine-tune MAE-Face for 20 epochs, with 10 warmup epochs. AdamW optimizer is used with a weight decay of 0.05. The batch size is 512. The learning rate is scheduled with cosine decay. RandAug(9, 0.5)\cite{RandAug} is applied as the data augmentation. For regularization, drop path\cite{DropPath} of 0.1 is used. Label smoothing\cite{LabelSmoothing} is not used because it degrades the performance. 
For AU detection, Mixup\cite{mixup} of 0.2 and cutmix\cite{cutmix} of 0.75 is used. We set the base learning rate to 1e-4, 2e-4, and 2e-4 for BP4D, BP4D+, and DISFA, respectively.
For AU intensity estimation, no mixup or cutmix is used. We set the base learning rate to 3e-5 and 1.5e-4 for BP4D and DISFA, respectively.

\begin{table*}[!t]

\setlength{\tabcolsep}{3.5pt}
\renewcommand{\arraystretch}{1.3}
\renewcommand{\extrarowheight}{0.5pt}

\caption{Comparison of AU intensity estimation results on BP4D and DISFA}
\label{table:results:au_int_all}
\centering

\begin{tabular}{ll|cccccc|ccccccccccccc}
    \hline
    &  & \multicolumn{6}{c|}{AU on BP4D dataset} & \multicolumn{13}{c}{AU on DISFA dataset} \\
    Metric & Method & 6 & 10 & 12 & 14 & 17 & Avg. & 1 & 2 & 4 & 5 & 6 & 9 & 12 & 15 & 17 & 20 & 25 & 26 & Avg. \\
    \hline
    \hline
    ICC↑ & ISIR\cite{ISIR} & .79 & .80 & .86 & \textbf{.71} & .44 & .72  & - & - & - & - & - & - & - & - & - & - & - & - & - \\
    & VGP-AE\cite{VGP-AE} & - & - & - & - & - & - & .48 & .47 & .62 & .19 & .50 & .42 & .80 & .19 & .36 & .15 & .84 & .53 & .46 \\
    & 2DC\cite{2DC} & - & - & - & - & - & - & .70 & .55 & .69 & .05 & .59 & .57 & \textbf{.88} & .32 & .10 & .08 & .90 & .50 & .50 \\
    & G2RL\cite{G2RL} & - & - & - & - & - & - & .71 & .31 & \textbf{.82} & .06 & .48 & \textbf{.67} & .68 & .21 & .47 & .17 & \textbf{.95} & .75 & .52 \\
    & RE-Net\cite{RE-Net} & - & - & - & - & - & - & .59 & .63 & .73 & \textbf{.82} & .49 & .50 & .73 & .29 & .21 & .03 & .90 & .60 & .54 \\
    & BORMIR\cite{BORMIR} & .73 & .68 & .86 & .37 & .47 & .62 & .20 & .25 & .30 & .17 & .39 & .18 & .58 & .16 & .23 & .09 & .71 & .15 & .28 \\
    & CCNN-IT\cite{CCNN-IT} & .75 & .69 & .86 & .40 & .45 & .63 & .20 & .12 & .46 & .08 & .48 & .44 & .73 & .29 & .45 & .21 & .60 & .46 & .38 \\
    & KJRE\cite{KJRE} & .71 & .61 & .87 & .39 & .42 & .60 & .27 & .35 & .25 & .33 & .51 & .31 & .67 & .14 & .17 & .20 & .74 & .25 & .35 \\
    & KBSS\cite{KBSS} & .76 & .75 & .85 & .49 & .51 & .67 & .23 & .11 & .48 & .25 & .50 & .25 & .71 & .22 & .25 & .06 & .83 & .41 & .36 \\
    & SCC-Heatmap\cite{SCC-Heatmap} & .74 & \textbf{.82} & .86 & .68 & .51 & .72 & \textbf{.73} & .44 & .74 & .06 & .27 & .51 & .71 & .04 & .37 & .04 & .94 & \textbf{.78} & .47 \\
    & HR\cite{HR} & \textbf{.82} & \textbf{.82} & .80 & \textbf{.71} & .50 & .73 & .56 & .52 & .75 & .42 & .51 & .55 & .82 & \textbf{.55} & .37 & .21 & .93 & .62 & .57 \\
    & APs\cite{APs} & \textbf{.82} & .80 & .86 & .69 & .51 & \textbf{.74} & .35 & .19 & .78 & .73 & .52 & .65 & .81 & .49 & \textbf{.61} & .28 & .92 & .67 & .58 \\
    & FRL\cite{FRL} & - & - & - & - & - & .719  & - & - & - & - & - & - & - & - & - & - & - & - & .598 \\
    & \textbf{MAE-Face} & \textbf{.818} & .743 & \textbf{.890} & .519 & \textbf{.731} & \textbf{.740}   & \textbf{.734} & \textbf{.658} & .764 & .686 & \textbf{.654} & .599 & .870 & .521 & .582 & \textbf{.313} & \textbf{.952} & .753 & \textbf{.674} \\
    \hline
    MSE↓ & ISIR\cite{ISIR} & .83 & \textbf{.80} & .62 & 1.14 & .84 & .85  & - & - & - & - & - & - & - & - & - & - & - & - & - \\
    & VGP-AE\cite{VGP-AE} & - & - & - & - & - & - & .51 & .32 & 1.13 & .08 & .56 & .31 & .47 & .20 & .28 & .16 & .49 & .44 & .41 \\
    & 2DC\cite{2DC} & - & - & - & - & - & - & .32 & .39 & .53 & .26 & .43 & .30 & .25 & .27 & .61 & .18 & .37 & .55 & .37 \\
    & HR\cite{HR} & .68 & \textbf{.80} & .79 & \textbf{.98} & .64 & .78 & .41 & .37 & .70 & .08 & .44 & .30 & .29 & .14 & .26 & .16 & .24 & .39 & .32 \\
    & APs\cite{APs} & .72 & .84 & .60 & 1.13 & .57 & \textbf{.77} & .68 & .59 & \textbf{.40} & \textbf{.03} & .49 & \textbf{.15} & .26 & .13 & .22 & .20 & .35 & \textbf{.17} & .30 \\
    & \textbf{MAE-Face} & \textbf{.652} & .943 & \textbf{.521} & 1.293 & \textbf{.482} & .778   & \textbf{.215} & \textbf{.198} & .420 & .061 & \textbf{.318} & .191 & \textbf{.226} & \textbf{.110} & \textbf{.176} & \textbf{.154} & \textbf{.152} & .247 & \textbf{.206} \\
    \hline
    MAE↓ & BORMIR\cite{BORMIR} & .85 & .90 & .68 & 1.05 & .79 & .85 & .88 & .78 & 1.24 & .59 & .77 & .78 & .76 & .56 & .72 & .63 & .90 & .88 & .79 \\
    & CCNN-IT\cite{CCNN-IT} & 1.17 & 1.43 & .97 & 1.65 & 1.08 & 1.26 & .73 & .72 & 1.03 & .21 & .72 & .51 & .72 & .43 & .50 & .44 & 1.16 & .79 & .66 \\
    & KJRE\cite{KJRE} & .82 & .95 & .64 & 1.08 & .85 & .87 & 1.02 & .92 & 1.86 & .70 & .79 & .87 & .77 & .60 & .80 & .72 & .96 & .94 & .91 \\
    & KBSS\cite{KBSS} & .56 & .65 & .48 & .98 & .63 & .66 & .48 & .49 & .57 & .08 & .26 & .22 & .33 & .15 & .44 & .22 & .43 & .36 & .33 \\
    & SCC-Heatmap\cite{SCC-Heatmap} & .61 & \textbf{.56} & .52 & \textbf{.73} & .50 & .58 & .16 & .16 & \textbf{.27} & \textbf{.03} & \textbf{.25} & \textbf{.13} & .32 & .15 & .20 & .09 & .30 & .32 & .20 \\
    & \textbf{MAE-Face} & \textbf{.478} & .635 & \textbf{.416} & .821 & \textbf{.411} & \textbf{.552}   & \textbf{.113} & \textbf{.093} & .279 & \textbf{.030} & .256 & \textbf{.129} & \textbf{.213} & \textbf{.087} & \textbf{.151} & \textbf{.084} & \textbf{.179} & \textbf{.222} & \textbf{.153} \\
    \hline
\end{tabular}

\centering
~\\
Most previous works report results with two decimal places. We report three decimal places for more precise comparisons in our experiments.

\end{table*}

\subsection{Comparison Results}

We compare our MAE-Face with the state-of-the-art works on 5 experiments (Ex) as follows:
\begin{Exlist}
\item AU detection on BP4D, results are reported in Table~\ref{table:results:au_det_bp4d} with 3-fold cross validation.
\item AU detection on BP4D+, results are reported in Table~\ref{table:results:au_det_bp4d+} with 3-fold cross validation.
\item AU detection on DISFA, results are reported in Table~\ref{table:results:au_det_disfa} with 3-fold cross validation.
\item AU intensity estimation on BP4D, results are reported in Table~\ref{table:results:au_int_all} with FERA2015 test partition.
\item AU intensity estimation on DISFA, results are reported in Table~\ref{table:results:au_int_all} with 3-fold cross validation.
\end{Exlist}

\textbf{AU detection}.
Observing Table~\ref{table:results:au_det_bp4d}, MAE-Face achieves the average F1 score of 67.4\% on BP4D, outperforming all the previous works. In Table~\ref{table:results:au_det_bp4d+} and Table~\ref{table:results:au_det_disfa}, MAE-Face, with the average F1 scores of 64.8\% on BP4D+ and 70.8\% on DISFA, also shows absolute superiority in comparison with all the previous works. It's worth noting that our MAE-Face exceeds the previous best F1 scores on BP4D, BP4D+, and DISFA by 2.5\%, 3.3\%, and 2.6\% respectively.
These results strongly prove the effectiveness and generality of the proposed method.

\textbf{AU intensity estimation} (Table~\ref{table:results:au_int_all}).
The results are evaluated with three metrics discussed in Section~\ref{section:experiments} including ICC, mean squared error (MSE), and mean absolute error (MAE). Note that not all of the previous works report the results on both BP4D and DISFA with all three metrics.
In terms of mean squared error and mean absolute error, MAE-Face outperforms all the previous works, proving its superiority.
If measured with ICC, our proposed MAE-Face achieves 0.740 on BP4D, which is the same as the previous best result of APs\cite{APs}. MAE-Face also achieves the state-of-the-art level of 0.674 on DISFA, which exceeds the previous best result (0.598) by quite a large margin.

These results suggest that, a more difficult dataset, such as DISFA, benefits more from the MAE-Face pre-training, which implies the potential of the proposed model for more challenging tasks.
Furthermore, the ICC of MAE-Face on each AU is rather close to the corresponding previous best from various methods. It indicates the big improvement of average ICC performance on MAE-Face is because it has nearly no short board on all the AUs nor any bias on AUs.
It shows that MAE-Face is insensitive to the imbalanced label distribution in the training set, which can overcome the overfitting caused by data scarcity.

\begin{figure}[!t]
    \centering
    \includegraphics[width=8.5cm]{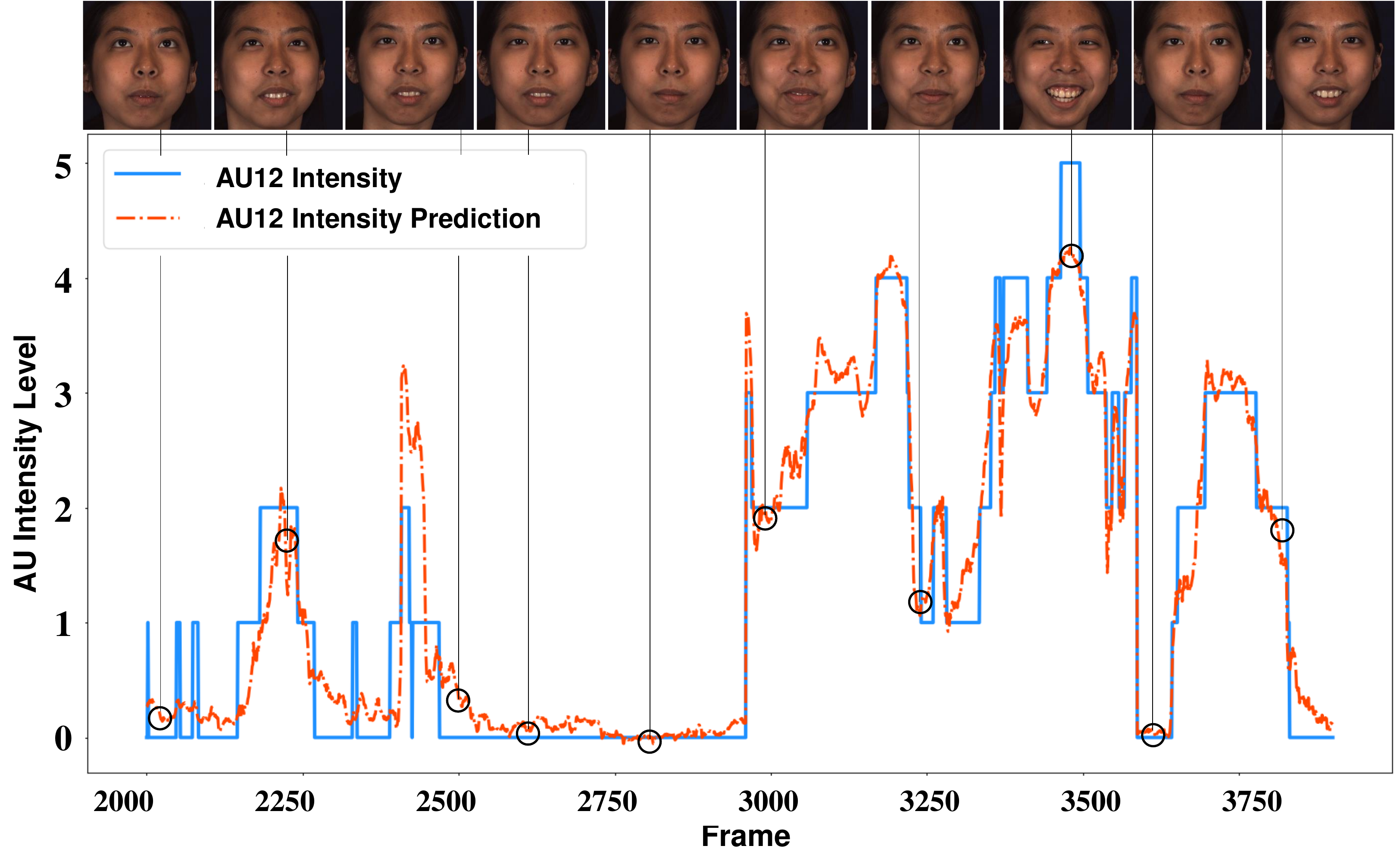}
    \caption{The sequence sample of AU12 intensity ground-truth and prediction of Subject F002 in BP4D test partition.}
    \label{fig:au_int_curve}
\end{figure}


Fig.~\ref{fig:au_int_curve} compares the predicted curve of MAE-Face with the ground-truth for AU12 intensity. This example comes from a video of Subject F002 in the BP4D test partition. We observe that the predicted curve sticks to the ground-truth values, which illustrates the high accuracy of our method in a direct view.

\subsection{Fine-tuning with Partial Dataset}
\label{section:results:finetune_partial_dataset}

\begin{figure}[!t]
    \centering
    \includegraphics[width=8.75cm]{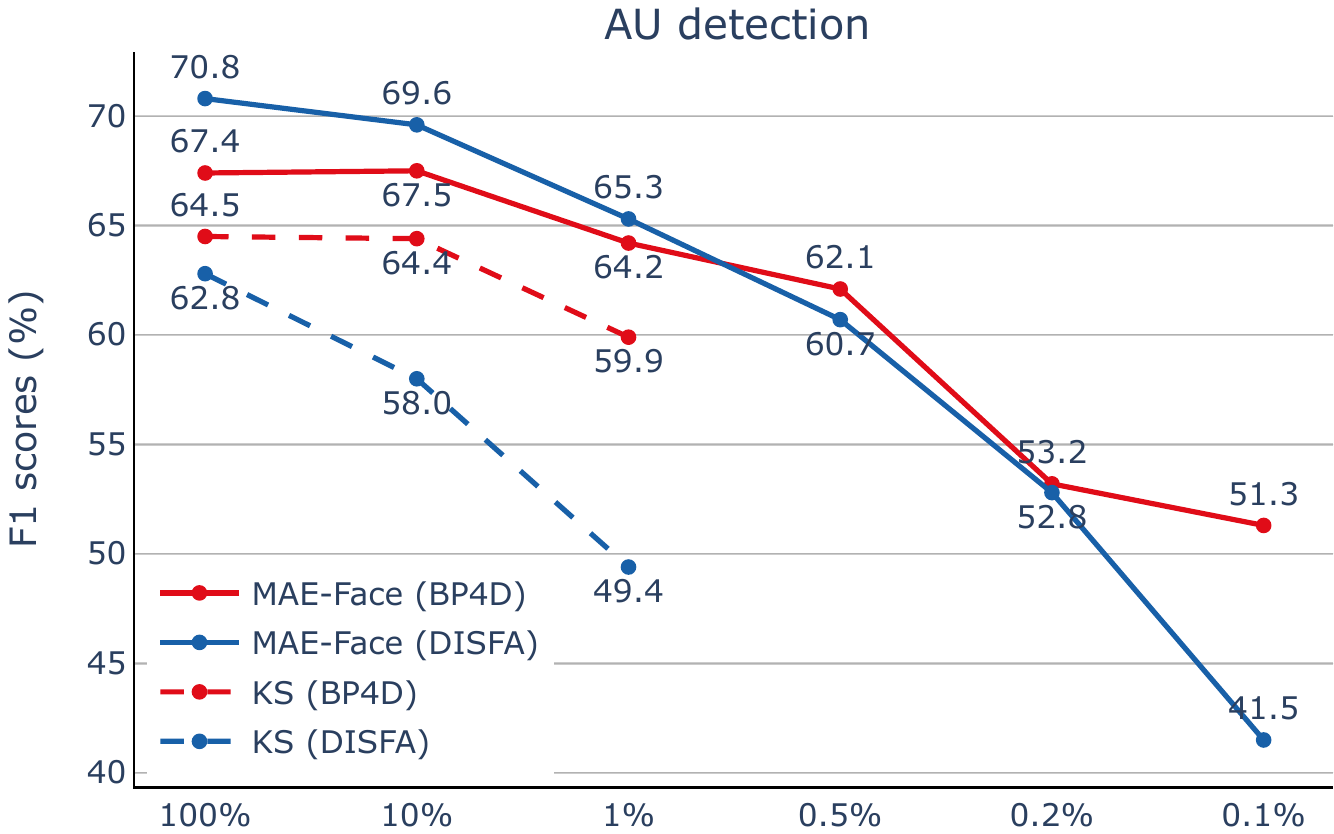}
    ~\\
    ~\\
    \includegraphics[width=8.75cm]{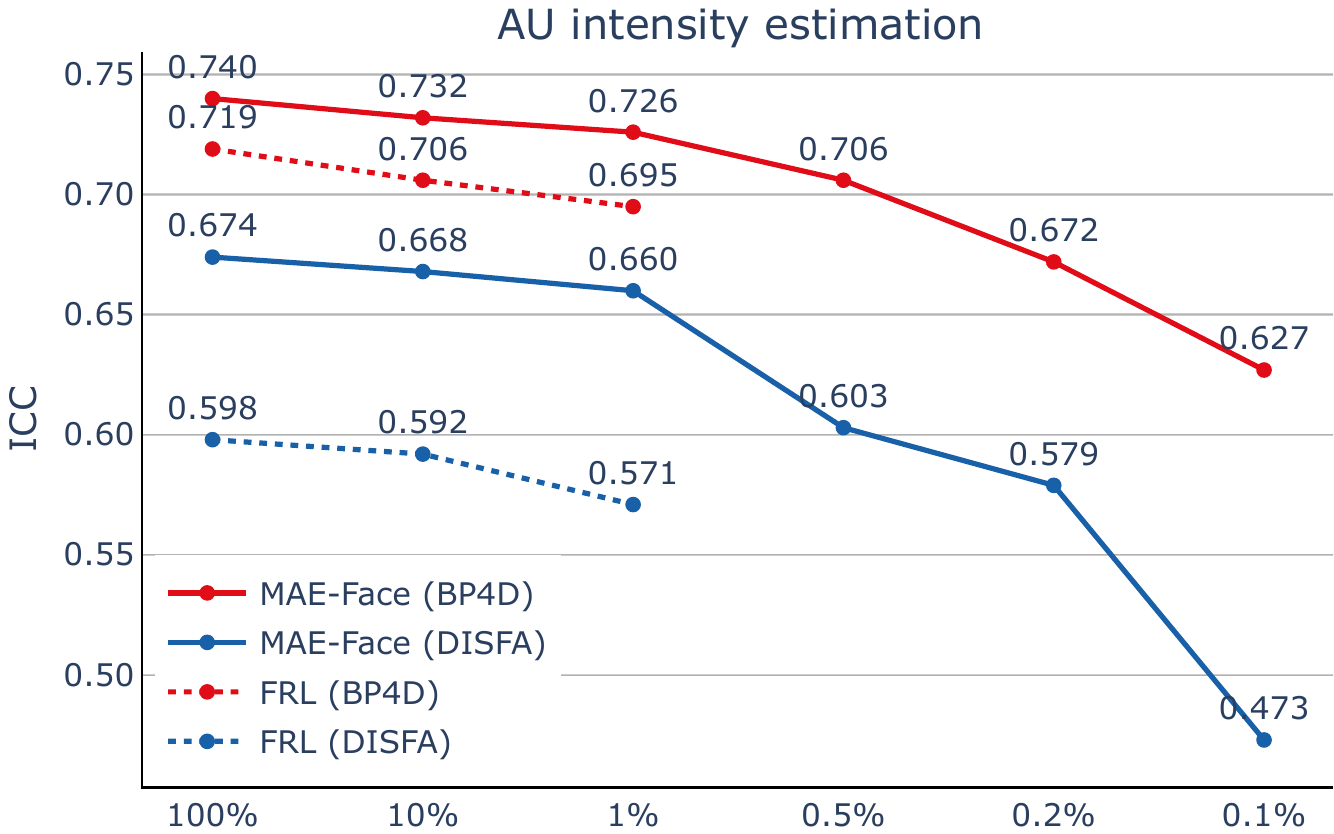}
    \caption{The results of fine-tuning on partial training set.}
    \label{fig:results:partial}
\end{figure}

To further dig into the generalization performance of MAE-Face, we carry out fine-tuning on a subset of the training data to observe if it can still perform well with a small amount of data, investigating its potential for few-shot learning.

We take a subset from the training set for training, with the test set unchanged for evaluation. The subset is built up by taking every one frame out of N frames. We fine-tune MAE-Face using 10\%, 1\%, 0.5\%, 0.2\% and 0.1\% of the training set, for 200, 2000, 4000, 10000 and 20000 epochs, respectively.

The results are shown in Fig.~\ref{fig:results:partial}, which also includes the results reported by KS\cite{KS} and FRL\cite{FRL} for comparison.
Interestingly, the performance of 10\% is almost the same as that of 100\% for most of the works. This is due to the temporal redundancy in video-based AU datasets, where you can take every 1 frame out of 10 frames and still retain most of the information.
When comparing 1\% to 10\% in terms of performance degradation, e.g. on DISFA, AU detection is 8.6\% and 4.3\% for KS and MAE-Face respectively, and AU intensity estimation is 0.021 and 0.008 for FRL and MAE-Face respectively.
Moreover, it's notable that the performance of MAE-Face on 0.5\% is still better than the performance of KS and FRL on 1\%.
We only observe significant performance degradation on MAE-Face when less than 0.5\% of the training set is used.

These results strongly prove the robustness of our proposed MAE-Face on small and sparse datasets, showing its potential to be a few-shot learner for action units and to solve real problems with very limited datasets.

\subsection{Ablation Studies}

\subsubsection{Pre-training model}
\label{section:ablation:pretrain_model}

\begin{table}[!t]

\renewcommand{\arraystretch}{1.3}
\renewcommand{\extrarowheight}{0.5pt}

\caption{Ablation study: pre-training model}
\label{table:ablation:pretrain_model}
\centering

\begin{tabular}{l|ccc|cc}
    \hline
    & \multicolumn{3}{c|}{AU detection (F1)} & \multicolumn{2}{c}{AU intensity (ICC)} \\
    Model & BP4D & BP4D+ & DISFA & BP4D & DISFA \\
    \hline
    \hline
    train from scratch & 50.0 & 54.3 & 49.1 & .567 & .418 \\
    MAE-IN1k & 63.3 & 61.9 & 67.6 & .694 & .599 \\
    \textbf{MAE-Face} & \textbf{67.4} & \textbf{64.8} & \textbf{70.8} & \textbf{.740} & \textbf{.674} \\
    \hline
\end{tabular}

\end{table}

MAE-Face benefits a lot from the proposed facial representation pre-training method. We conduct further experiments to confirm
\begin{inlinelist}
\item whether MAE-Face benefits from the two-stage pre-training framework, and
\item whether MAE-Face benefits from pre-training on face images instead of general images.
\end{inlinelist}

Table~\ref{table:ablation:pretrain_model} shows the results of the one-stage training (\textit{train from scratch}), the pre-training using general images (MAE-IN1k), and the pre-training using face images (the proposed MAE-Face).
\begin{inlinelist}
\item \textit{train from scratch} refers to training from scratch using randomly initialized weights on AU datasets, where the training takes 200 epochs with 20 warmup epochs.
\item MAE-IN1k refers to fine-tuned from the ImageNet-1k\cite{ImageNet} pre-trained MAE\cite{MAE}, where we use the same fine-tuning settings as that of MAE-Face.
\item MAE-Face refers to fine-tuned from the proposed pre-training dataset.
\end{inlinelist}
All the above models use the same ViT-Base backbone and the same regularization tricks.
The major difference is that they are initialized with different weights for the training of AU-related tasks.

As the results show in Table~\ref{table:ablation:pretrain_model}, \textit{train from scratch} has the worst performance when compared to the others.
By investigating its training progress, we find that its convergence is slow, yet it still overfits the training set easily.
MAE-IN1k improves the performance significantly in comparison to \textit{train from scratch}.
MAE-Face further improves the results by a large margin compared to MAE-IN1k and \textit{train from scratch}.
This result proves that a good initialization is a key point for an AU analysis model to converge and generalize, and our facial representation model MAE-Face serves as a pretty good one.
These results confirm the effectiveness of the MAE-Face method, rather than its backbone or the regularization tricks.

\subsubsection{Pre-training loss function}
\label{section:ablation:pretrain_loss_function}

\begin{figure}[!t]
    \centering
    \includegraphics[width=9cm]{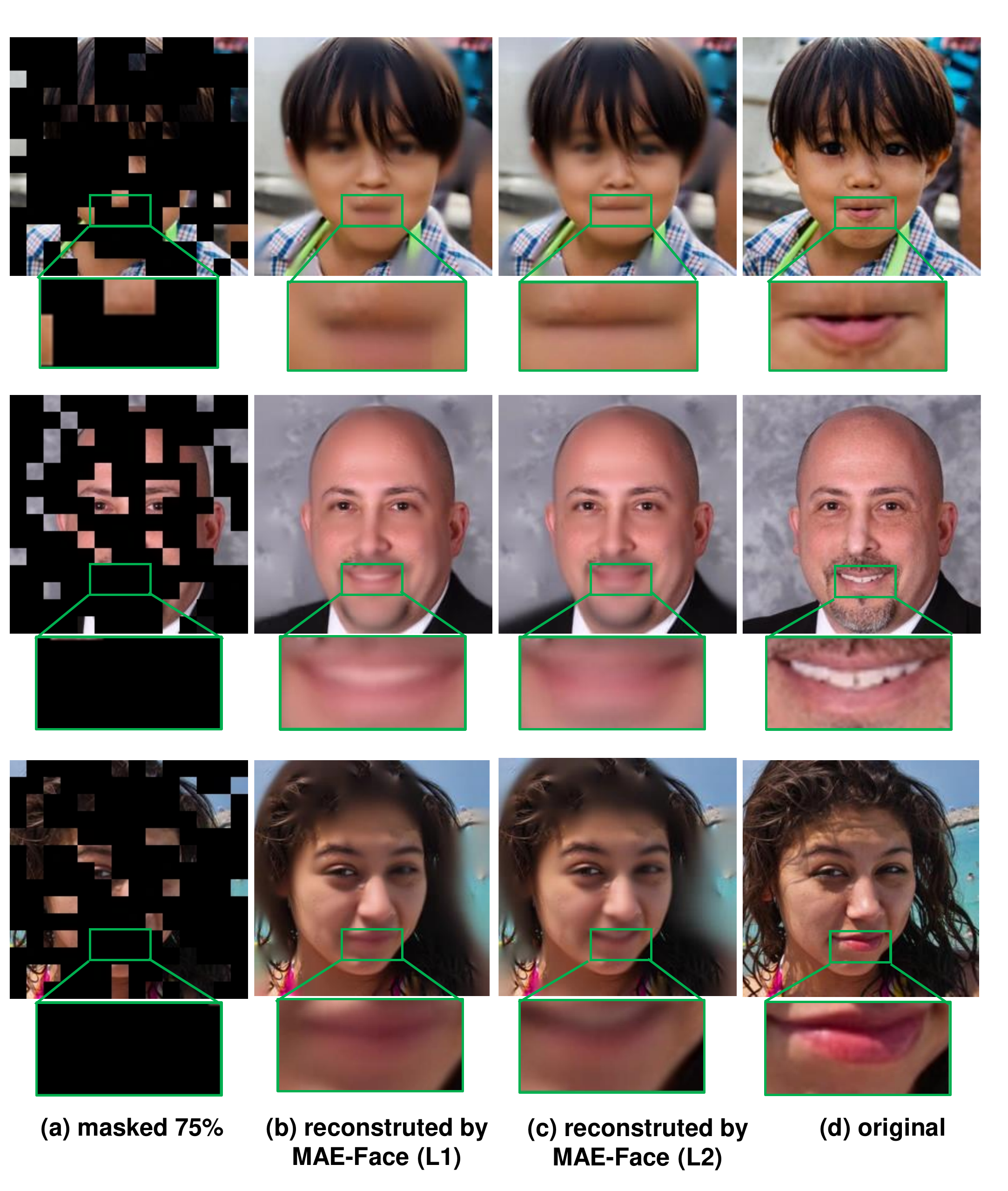}
    \caption{Reconstructed samples on FFHQ images, using MAE-Face trained with L1 loss or L2 loss.}
    \label{fig:ablation:loss}
\end{figure}

\begin{table}[!t]

\renewcommand{\arraystretch}{1.3}
\renewcommand{\extrarowheight}{0.5pt}

\caption{Ablation study: loss function for pre-training}
\label{table:ablation:loss}
\centering

\begin{tabular}{l|ccc|cc}
    \hline
    & \multicolumn{3}{c|}{AU detection (F1)} & \multicolumn{2}{c}{AU intensity (ICC)} \\
    Loss function & BP4D & BP4D+ & DISFA & BP4D & DISFA \\
    \hline
    \hline
    L2 (w/o norm) & 64.7 & 64.4 & 69.3 & .710 & .625 \\
    L2 (w/ norm) & 66.4 & 64.4 & 69.8 & .731 & .657 \\
    L1 (w/o norm) & 66.3 & 63.9 & 68.9 & .728 & .646 \\
    \textbf{L1 (w/ norm)} & \textbf{66.9} & \textbf{64.6} & \textbf{70.8} & \textbf{.736} & \textbf{.665} \\
    \hline
\end{tabular}

\centering
~\\
The tested models in this table are pre-trained for 200 epochs.

\end{table}

Our proposed MAE-Face uses L1 loss instead of L2 loss to regress the reconstructed pixels.
In Table~\ref{table:ablation:loss}, we test the impact of different loss functions for pre-training.
We also test the impact of patch-wise normalization (\textit{w/ norm}), which has also been studied in\cite{MAE}.
The experimental results confirm the effectiveness of both L1 loss and patch-wise normalization.

Specifically, we present some reconstructed samples of L1 loss or L2 loss in Fig.~\ref{fig:ablation:loss}.
We observe that the facial expression reconstructed by L1 loss is more accurate than L2 loss, especially in the mouth part.
It proves that the MAE-Face trained with L1 loss learns a better facial expression representation.

As illustrated by \cite{zhao2015loss}, L2 loss tends to over-penalize large errors while under-penalize small errors, usually performing worse than L1 loss in image restoration. Masked autoencoding is another image restoration problem, which is also expected to take advantage of L1 loss. Our experiments confirm this expectation.


\section{Discussion and Conclusion}
\label{section:conclusion}

This work describes a self-supervised pre-training framework specifically for face images, achieving really good results for AU detection and AU intensity estimation on available datasets.
The pre-trained model, named MAE-Face, benefits from the masked autoencoding paradigm to learn a robust facial representation.

By comparing the proposed MAE-Face with the previous works, MAE-Face has reached new state-of-the-arts on nearly all the evaluated subjects.
Furthermore, when fine-tuned on a subset of the training set, MAE-Face still exhibits very good results.
The performance degradation is subtle even when fine-tuned on 1\% of the training set, which shows its robustness to generalize on limited and biased data.
Since the AU annotation requires domain expertise, MAE-Face may greatly ease the efforts required for building an AU dataset in the future.

Besides, we carry out ablations to survey the performance of MAE-Face on different setups.
We verify that MAE-Face benefits from both masked autoencoding and the specific pre-training for facial representations.
We also verify that MAE-Face pre-training benefits from both L1 loss function and patch-wise normalization to boost its performance for downstream tasks.

We've also found some limitations of MAE-Face:
\begin{inlinelist}
\item It still relies on face alignment to get the optimal performance, which may bring extra computational burdens for deployment.
\item It benefits from a carefully tuned learning rate to get the optimal performance on each dataset.
\end{inlinelist}
Nevertheless, when searching for the optimal setups, even without face alignment and learning rate search, MAE-Face has already surpassed the previous bests on most of the tasks.

To this end, we conclude that our proposed MAE-Face is a strong learner for AU detection and AU intensity estimation.
However, since MAE-Face is pre-trained on face images without any specific designs for action units, perhaps its potential is far more than that.
Thus for future works, it is interesting to survey that,
\begin{inlinelist}
\item if it can adapt to other kinds of face-related tasks, and
\item if it can work well on much smaller backbone other than ViT-Base (e.g. through distillation).
\end{inlinelist}
With which, it has the potential to become a universal facial representation model for a bunch of face-related tasks, as well as a lightweight model for easy deployments.

\appendices




\ifCLASSOPTIONcompsoc
  \section*{Acknowledgments}
\else
  \section*{Acknowledgment}
\fi

This research is supported by the Key Research and Development Program of Zhejiang Province (No. 2022C01011) and the 2022 Key Artificial Intelligence Science and Technology Innovation Project of Hangzhou Science and Technology Office.

\ifCLASSOPTIONcaptionsoff
  \newpage
\fi





\bibliographystyle{ieeetr} 
\bibliography{egbib} 

\begin{thebibliography}{100}

\bibitem{FACS}
P.~Ekman and W.~Friesen, ``Facial action coding system: A technique for the
  measurement of facial movement,'' {\em Consulting Psychologists Press Palo
  Alto}, vol.~12, 01 1978.

\bibitem{lucey2010automatically}
P.~Lucey, J.~F. Cohn, I.~Matthews, S.~Lucey, S.~Sridharan, J.~Howlett, and
  K.~M. Prkachin, ``Automatically detecting pain in video through facial action
  units,'' {\em IEEE Transactions on Systems, Man, and Cybernetics, Part B
  (Cybernetics)}, vol.~41, no.~3, pp.~664--674, 2010.

\bibitem{bevilacqua2016variations}
F.~Bevilacqua, P.~Backlund, and H.~Engstrom, ``Variations of facial actions
  while playing games with inducing boredom and stress,'' in {\em 2016 8th
  International Conference on Games and Virtual Worlds for Serious Applications
  (VS-GAMES)}, pp.~1--8, IEEE, 2016.

\bibitem{sikander2020novel}
G.~Sikander and S.~Anwar, ``A novel machine vision-based 3d facial action unit
  identification for fatigue detection,'' {\em IEEE Transactions on Intelligent
  Transportation Systems}, vol.~22, no.~5, pp.~2730--2740, 2020.

\bibitem{yan2022weakly}
J.~Yan, J.~Wang, Q.~Li, C.~Wang, and S.~Pu, ``Weakly supervised regional and
  temporal learning for facial action unit recognition,'' {\em IEEE
  Transactions on Multimedia}, 2022.

\bibitem{niu2019multi}
X.~Niu, H.~Han, S.~Shan, and X.~Chen, ``Multi-label co-regularization for
  semi-supervised facial action unit recognition,'' {\em Advances in neural
  information processing systems}, vol.~32, 2019.

\bibitem{chang2022knowledge}
Y.~Chang and S.~Wang, ``Knowledge-driven self-supervised representation
  learning for facial action unit recognition,'' in {\em Proceedings of the
  IEEE/CVF Conference on Computer Vision and Pattern Recognition},
  pp.~20417--20426, 2022.

\bibitem{zhang2021prior}
W.~Zhang, Z.~Guo, K.~Chen, L.~Li, Z.~Zhang, Y.~Ding, R.~Wu, T.~Lv, and C.~Fan,
  ``Prior aided streaming network for multi-task affective analysis,'' in {\em
  Proceedings of the IEEE/CVF International Conference on Computer Vision},
  pp.~3539--3549, 2021.

\bibitem{chen2021understanding}
Y.~Chen and J.~Joo, ``Understanding and mitigating annotation bias in facial
  expression recognition,'' in {\em Proceedings of the IEEE/CVF International
  Conference on Computer Vision}, pp.~14980--14991, 2021.

\bibitem{BERT}
J.~Devlin, M.-W. Chang, K.~Lee, and K.~Toutanova, ``Bert: Pre-training of deep
  bidirectional transformers for language understanding,'' {\em arXiv preprint
  arXiv:1810.04805}, 2018.

\bibitem{GPT-3}
T.~Brown, B.~Mann, N.~Ryder, M.~Subbiah, J.~D. Kaplan, P.~Dhariwal,
  A.~Neelakantan, P.~Shyam, G.~Sastry, A.~Askell, {\em et~al.}, ``Language
  models are few-shot learners,'' {\em Advances in neural information
  processing systems}, vol.~33, pp.~1877--1901, 2020.

\bibitem{BEiT}
H.~Bao, L.~Dong, and F.~Wei, ``Beit: Bert pre-training of image transformers,''
  {\em arXiv preprint arXiv:2106.08254}, 2021.

\bibitem{MAE}
K.~He, X.~Chen, S.~Xie, Y.~Li, P.~Doll{\'a}r, and R.~Girshick, ``Masked
  autoencoders are scalable vision learners,'' in {\em Proceedings of the
  IEEE/CVF Conference on Computer Vision and Pattern Recognition},
  pp.~16000--16009, 2022.

\bibitem{ImageNet}
O.~Russakovsky, J.~Deng, H.~Su, J.~Krause, S.~Satheesh, S.~Ma, Z.~Huang,
  A.~Karpathy, A.~Khosla, M.~Bernstein, {\em et~al.}, ``Imagenet large scale
  visual recognition challenge,'' {\em International journal of computer
  vision}, vol.~115, no.~3, pp.~211--252, 2015.

\bibitem{valstar2006fully}
M.~Valstar and M.~Pantic, ``Fully automatic facial action unit detection and
  temporal analysis,'' in {\em 2006 Conference on Computer Vision and Pattern
  Recognition Workshop (CVPRW'06)}, pp.~149--149, IEEE, 2006.

\bibitem{jiang2011action}
B.~Jiang, M.~F. Valstar, and M.~Pantic, ``Action unit detection using sparse
  appearance descriptors in space-time video volumes,'' in {\em 2011 IEEE
  International Conference on Automatic Face \& Gesture Recognition (FG)},
  pp.~314--321, IEEE, 2011.

\bibitem{LP48}
L.~Zhong, Q.~Liu, P.~Yang, J.~Huang, and D.~Metaxas, ``Learning multiscale
  active facial patches for expression analysis,'' {\em IEEE Transactions on
  Cybernetics}, vol.~45, pp.~1499--1510, 8 2015.

\bibitem{zeng2015confidence}
J.~Zeng, W.-S. Chu, F.~De~la Torre, J.~F. Cohn, and Z.~Xiong, ``Confidence
  preserving machine for facial action unit detection,'' in {\em Proceedings of
  the IEEE international conference on computer vision}, pp.~3622--3630, 2015.

\bibitem{Onal2019D-PAttNet}
I.~Onal~Ertugrul, L.~Yang, L.~A. Jeni, and J.~F. Cohn, ``D-pattnet: Dynamic
  patch-attentive deep network for action unit detection,'' {\em Frontiers in
  computer science}, vol.~1, p.~11, 2019.

\bibitem{JPML}
K.~{Zhao}, W.~S. {Chu}, F.~{De la Torre}, J.~F. {Cohn}, and H.~{Zhang}, ``Joint
  patch and multi-label learning for facial action unit and holistic expression
  recognition,'' {\em IEEE Transactions on Image Processing}, vol.~25, no.~8,
  pp.~3931--3946, 2016.

\bibitem{DRML}
K.~Zhao, W.-S. Chu, and H.~Zhang, ``Deep region and multi-label learning for
  facial action unit detection,'' in {\em Proceedings of the IEEE conference on
  computer vision and pattern recognition}, pp.~3391--3399, 2016.

\bibitem{ma2019r}
C.~Ma, L.~Chen, and J.~Yong, ``Au r-cnn: Encoding expert prior knowledge into
  r-cnn for action unit detection,'' {\em neurocomputing}, vol.~355,
  pp.~35--47, 2019.

\bibitem{ge2021local}
X.~Ge, P.~Wan, H.~Han, J.~M. Jose, Z.~Ji, Z.~Wu, and X.~Liu, ``Local global
  relational network for facial action units recognition,'' in {\em 2021 16th
  IEEE International Conference on Automatic Face and Gesture Recognition (FG
  2021)}, pp.~01--08, IEEE, 2021.

\bibitem{ge2022mgrr}
X.~Ge, J.~M. Jose, S.~Xu, X.~Liu, and H.~Han, ``Mgrr-net: Multi-level graph
  relational reasoning network for facial action units detection,'' {\em arXiv
  preprint arXiv:2204.01349}, 2022.

\bibitem{tong2007facial}
Y.~Tong, W.~Liao, and Q.~Ji, ``Facial action unit recognition by exploiting
  their dynamic and semantic relationships,'' {\em IEEE transactions on pattern
  analysis and machine intelligence}, vol.~29, no.~10, pp.~1683--1699, 2007.

\bibitem{LP}
X.~Niu, H.~Han, S.~Yang, Y.~Huang, and S.~Shan, ``Local relationship learning
  with person-specific shape regularization for facial action unit detection,''
  in {\em Proceedings of the IEEE Conference on Computer Vision and Pattern
  Recognition}, pp.~11917--11926, 2019.

\bibitem{corneanu2018deep}
C.~Corneanu, M.~Madadi, and S.~Escalera, ``Deep structure inference network for
  facial action unit recognition,'' in {\em Proceedings of the european
  conference on computer vision (ECCV)}, pp.~298--313, 2018.

\bibitem{liu2020relation}
Z.~Liu, J.~Dong, C.~Zhang, L.~Wang, and J.~Dang, ``Relation modeling with graph
  convolutional networks for facial action unit detection,'' in {\em
  International Conference on Multimedia Modeling}, pp.~489--501, Springer,
  2020.

\bibitem{EAC-Net}
W.~Li, F.~Abtahi, Z.~Zhu, and L.~Yin, ``Eac-net: A region-based deep enhancing
  and cropping approach for facial action unit detection,'' in {\em 2017 12th
  IEEE International Conference on Automatic Face \& Gesture Recognition (FG
  2017)}, pp.~103--110, IEEE, 2017.

\bibitem{shao2021jaa}
Z.~Shao, Z.~Liu, J.~Cai, and L.~Ma, ``Jaa-net: joint facial action unit
  detection and face alignment via adaptive attention,'' {\em International
  Journal of Computer Vision}, vol.~129, no.~2, pp.~321--340, 2021.

\bibitem{attnetionTrans2019}
Z.~Shao, Z.~Liu, J.~Cai, Y.~Wu, and L.~Ma, ``Facial action unit detection using
  attention and relation learning,'' {\em IEEE Transactions on Affective
  Computing}, 2019.

\bibitem{jacob2021facial}
G.~M. Jacob and B.~Stenger, ``Facial action unit detection with transformers,''
  in {\em Proceedings of the IEEE/CVF Conference on Computer Vision and Pattern
  Recognition}, pp.~7680--7689, 2021.

\bibitem{cui2020knowledge}
Z.~Cui, T.~Song, Y.~Wang, and Q.~Ji, ``Knowledge augmented deep neural networks
  for joint facial expression and action unit recognition,'' {\em Advances in
  Neural Information Processing Systems}, vol.~33, 2020.

\bibitem{yang2021exploiting}
H.~Yang, L.~Yin, Y.~Zhou, and J.~Gu, ``Exploiting semantic embedding and visual
  feature for facial action unit detection,'' in {\em Proceedings of the
  IEEE/CVF Conference on Computer Vision and Pattern Recognition},
  pp.~10482--10491, 2021.

\bibitem{zhao2018learning}
K.~Zhao, W.-S. Chu, and A.~M. Martinez, ``Learning facial action units from web
  images with scalable weakly supervised clustering,'' in {\em Proceedings of
  the IEEE Conference on computer vision and pattern recognition},
  pp.~2090--2099, 2018.

\bibitem{zhang2022transformer}
W.~Zhang, F.~Qiu, S.~Wang, H.~Zeng, Z.~Zhang, R.~An, B.~Ma, and Y.~Ding,
  ``Transformer-based multimodal information fusion for facial expression
  analysis,'' in {\em Proceedings of the IEEE/CVF Conference on Computer Vision
  and Pattern Recognition}, pp.~2428--2437, 2022.

\bibitem{zhang2021learning}
W.~Zhang, X.~Ji, K.~Chen, Y.~Ding, and C.~Fan, ``Learning a facial expression
  embedding disentangled from identity,'' in {\em Proceedings of the IEEE/CVF
  Conference on Computer Vision and Pattern Recognition}, pp.~6759--6768, 2021.

\bibitem{chen2022causal}
Y.~Chen, D.~Chen, T.~Wang, Y.~Wang, and Y.~Liang, ``Causal intervention for
  subject-deconfounded facial action unit recognition,'' {\em arXiv preprint
  arXiv:2204.07935}, 2022.

\bibitem{KS}
X.~Li, X.~Zhang, T.~Wang, and L.~Yin, ``Knowledge-spreader: Learning facial
  action unit dynamics with extremely limited labels,'' {\em arXiv preprint
  arXiv:2203.16678}, 2022.

\bibitem{rudovic2012multi}
O.~Rudovic, V.~Pavlovic, and M.~Pantic, ``Multi-output laplacian dynamic
  ordinal regression for facial expression recognition and intensity
  estimation,'' in {\em 2012 IEEE Conference on Computer Vision and Pattern
  Recognition}, pp.~2634--2641, IEEE, 2012.

\bibitem{rudovic2012kernel}
O.~Rudovic, V.~Pavlovic, and M.~Pantic, ``Kernel conditional ordinal random
  fields for temporal segmentation of facial action units,'' in {\em European
  Conference on Computer Vision}, pp.~260--269, Springer, 2012.

\bibitem{eleftheriadis2017gaussian}
S.~Eleftheriadis, O.~Rudovic, M.~P. Deisenroth, and M.~Pantic, ``Gaussian
  process domain experts for modeling of facial affect,'' {\em IEEE
  transactions on image processing}, vol.~26, no.~10, pp.~4697--4711, 2017.

\bibitem{CCNN-IT}
R.~Walecki, V.~Pavlovic, B.~Schuller, M.~Pantic, {\em et~al.}, ``Deep
  structured learning for facial action unit intensity estimation,'' in {\em
  Proceedings of the IEEE Conference on Computer Vision and Pattern
  Recognition}, pp.~3405--3414, 2017.

\bibitem{VGP-AE}
S.~Eleftheriadis, O.~Rudovic, M.~P. Deisenroth, and M.~Pantic, ``Variational
  gaussian process auto-encoder for ordinal prediction of facial action
  units,'' in {\em Asian Conference on Computer Vision}, pp.~154--170,
  Springer, 2016.

\bibitem{sandbach2013markov}
G.~Sandbach, S.~Zafeiriou, and M.~Pantic, ``Markov random field structures for
  facial action unit intensity estimation,'' in {\em Proceedings of the IEEE
  International Conference on Computer Vision Workshops}, pp.~738--745, 2013.

\bibitem{KBSS}
Y.~Zhang, W.~Dong, B.-G. Hu, and Q.~Ji, ``Weakly-supervised deep convolutional
  neural network learning for facial action unit intensity estimation,'' in
  {\em Proceedings of the IEEE Conference on Computer Vision and Pattern
  Recognition}, pp.~2314--2323, 2018.

\bibitem{BORMIR}
Y.~Zhang, R.~Zhao, W.~Dong, B.-G. Hu, and Q.~Ji, ``Bilateral ordinal relevance
  multi-instance regression for facial action unit intensity estimation,'' in
  {\em Proceedings of the IEEE conference on computer vision and pattern
  recognition}, pp.~7034--7043, 2018.

\bibitem{KJRE}
Y.~Zhang, B.~Wu, W.~Dong, Z.~Li, W.~Liu, B.-G. Hu, and Q.~Ji, ``Joint
  representation and estimator learning for facial action unit intensity
  estimation,'' in {\em Proceedings of the IEEE/CVF Conference on Computer
  Vision and Pattern Recognition}, pp.~3457--3466, 2019.

\bibitem{HR}
I.~Ntinou, E.~Sanchez, A.~Bulat, M.~Valstar, and Y.~Tzimiropoulos, ``A transfer
  learning approach to heatmap regression for action unit intensity
  estimation,'' {\em IEEE Transactions on Affective Computing}, 2021.

\bibitem{SCC-Heatmap}
Y.~Fan, J.~Lam, and V.~Li, ``Facial action unit intensity estimation via
  semantic correspondence learning with dynamic graph convolution,'' in {\em
  Proceedings of the AAAI Conference on Artificial Intelligence}, vol.~34,
  pp.~12701--12708, 2020.

\bibitem{zhang2019context}
Y.~Zhang, H.~Jiang, B.~Wu, Y.~Fan, and Q.~Ji, ``Context-aware feature and label
  fusion for facial action unit intensity estimation with partially labeled
  data,'' in {\em Proceedings of the IEEE/CVF International Conference on
  Computer Vision}, pp.~733--742, 2019.

\bibitem{APs}
E.~Sanchez, M.~K. Tellamekala, M.~Valstar, and G.~Tzimiropoulos, ``Affective
  processes: stochastic modelling of temporal context for emotion and facial
  expression recognition,'' in {\em Proceedings of the IEEE/CVF Conference on
  Computer Vision and Pattern Recognition}, pp.~9074--9084, 2021.

\bibitem{wu2018unsupervised}
Z.~Wu, Y.~Xiong, S.~X. Yu, and D.~Lin, ``Unsupervised feature learning via
  non-parametric instance discrimination,'' in {\em Proceedings of the IEEE
  conference on computer vision and pattern recognition}, pp.~3733--3742, 2018.

\bibitem{oord2018representation}
A.~v.~d. Oord, Y.~Li, and O.~Vinyals, ``Representation learning with
  contrastive predictive coding,'' {\em arXiv preprint arXiv:1807.03748}, 2018.

\bibitem{hjelm2018learning}
R.~D. Hjelm, A.~Fedorov, S.~Lavoie-Marchildon, K.~Grewal, P.~Bachman,
  A.~Trischler, and Y.~Bengio, ``Learning deep representations by mutual
  information estimation and maximization,'' in {\em International Conference
  on Learning Representations}, 2019.

\bibitem{bachman2019learning}
P.~Bachman, R.~D. Hjelm, and W.~Buchwalter, ``Learning representations by
  maximizing mutual information across views,'' {\em Advances in neural
  information processing systems}, vol.~32, 2019.

\bibitem{he2020momentum}
K.~He, H.~Fan, Y.~Wu, S.~Xie, and R.~Girshick, ``Momentum contrast for
  unsupervised visual representation learning,'' in {\em Proceedings of the
  IEEE/CVF conference on computer vision and pattern recognition},
  pp.~9729--9738, 2020.

\bibitem{chen2020simple}
T.~Chen, S.~Kornblith, M.~Norouzi, and G.~Hinton, ``A simple framework for
  contrastive learning of visual representations,'' in {\em International
  conference on machine learning}, pp.~1597--1607, PMLR, 2020.

\bibitem{ViT}
A.~Dosovitskiy, L.~Beyer, A.~Kolesnikov, D.~Weissenborn, X.~Zhai,
  T.~Unterthiner, M.~Dehghani, M.~Minderer, G.~Heigold, S.~Gelly, {\em et~al.},
  ``An image is worth 16x16 words: Transformers for image recognition at
  scale,'' {\em arXiv preprint arXiv:2010.11929}, 2020.

\bibitem{DALLE}
A.~Ramesh, M.~Pavlov, G.~Goh, S.~Gray, C.~Voss, A.~Radford, M.~Chen, and
  I.~Sutskever, ``Zero-shot text-to-image generation,'' in {\em International
  Conference on Machine Learning}, pp.~8821--8831, PMLR, 2021.

\bibitem{chen2021cafgraph}
Y.~Chen, D.~Chen, Y.~Wang, T.~Wang, and Y.~Liang, ``Cafgraph: Context-aware
  facial multi-graph representation for facial action unit recognition,'' in
  {\em Proceedings of the 29th ACM International Conference on Multimedia},
  pp.~1029--1037, 2021.

\bibitem{chou2020remix}
H.-P. Chou, S.-C. Chang, J.-Y. Pan, W.~Wei, and D.-C. Juan, ``Remix: rebalanced
  mixup,'' in {\em European Conference on Computer Vision}, pp.~95--110,
  Springer, 2020.

\bibitem{liu2019large}
Z.~Liu, Z.~Miao, X.~Zhan, J.~Wang, B.~Gong, and S.~X. Yu, ``Large-scale
  long-tailed recognition in an open world,'' in {\em Proceedings of the
  IEEE/CVF Conference on Computer Vision and Pattern Recognition},
  pp.~2537--2546, 2019.

\bibitem{cui2019class}
Y.~Cui, M.~Jia, T.-Y. Lin, Y.~Song, and S.~Belongie, ``Class-balanced loss
  based on effective number of samples,'' in {\em Proceedings of the IEEE/CVF
  conference on computer vision and pattern recognition}, pp.~9268--9277, 2019.

\bibitem{xiang2017linear}
X.~Xiang and T.~D. Tran, ``Linear disentangled representation learning for
  facial actions,'' {\em IEEE Transactions on Circuits and Systems for Video
  Technology}, vol.~28, no.~12, pp.~3539--3544, 2017.

\bibitem{li2020deep}
S.~Li and W.~Deng, ``Deep facial expression recognition: A survey,'' {\em IEEE
  transactions on affective computing}, 2020.

\bibitem{Transformer}
A.~Vaswani, N.~Shazeer, N.~Parmar, J.~Uszkoreit, L.~Jones, A.~N. Gomez,
  {\L}.~Kaiser, and I.~Polosukhin, ``Attention is all you need,'' {\em Advances
  in neural information processing systems}, vol.~30, 2017.

\bibitem{AffectNet}
A.~Mollahosseini, B.~Hasani, and M.~H. Mahoor, ``Affectnet: A database for
  facial expression, valence, and arousal computing in the wild,'' {\em IEEE
  Transactions on Affective Computing}, vol.~10, no.~1, pp.~18--31, 2017.

\bibitem{CASIA-WebFace}
D.~Yi, Z.~Lei, S.~Liao, and S.~Z. Li, ``Learning face representation from
  scratch,'' {\em arXiv preprint arXiv:1411.7923}, 2014.

\bibitem{IMDB-WIKI}
R.~Rothe, R.~Timofte, and L.~Van~Gool, ``Deep expectation of real and apparent
  age from a single image without facial landmarks,'' {\em International
  Journal of Computer Vision}, vol.~126, no.~2, pp.~144--157, 2018.

\bibitem{CelebA}
Z.~Liu, P.~Luo, X.~Wang, and X.~Tang, ``Deep learning face attributes in the
  wild,'' in {\em Proceedings of the IEEE international conference on computer
  vision}, pp.~3730--3738, 2015.

\bibitem{RetinaFace}
J.~Deng, J.~Guo, E.~Ververas, I.~Kotsia, and S.~Zafeiriou, ``Retinaface:
  Single-shot multi-level face localisation in the wild,'' in {\em Proceedings
  of the IEEE/CVF conference on computer vision and pattern recognition},
  pp.~5203--5212, 2020.

\bibitem{LayerNorm}
J.~L. Ba, J.~R. Kiros, and G.~E. Hinton, ``Layer normalization,'' {\em arXiv
  preprint arXiv:1607.06450}, 2016.

\bibitem{BP4D}
X.~Zhang, L.~Yin, J.~F. Cohn, S.~Canavan, M.~Reale, A.~Horowitz, P.~Liu, and
  J.~M. Girard, ``Bp4d-spontaneous: a high-resolution spontaneous 3d dynamic
  facial expression database,'' {\em Image and Vision Computing}, vol.~32,
  no.~10, pp.~692--706, 2014.

\bibitem{BP4D+}
Z.~Zhang, J.~M. Girard, Y.~Wu, X.~Zhang, P.~Liu, U.~Ciftci, S.~Canavan,
  M.~Reale, A.~Horowitz, H.~Yang, {\em et~al.}, ``Multimodal spontaneous
  emotion corpus for human behavior analysis,'' in {\em Proceedings of the IEEE
  conference on computer vision and pattern recognition}, pp.~3438--3446, 2016.

\bibitem{DISFA}
S.~M. Mavadati, M.~H. Mahoor, K.~Bartlett, P.~Trinh, and J.~F. Cohn, ``Disfa: A
  spontaneous facial action intensity database,'' {\em IEEE Transactions on
  Affective Computing}, vol.~4, no.~2, pp.~151--160, 2013.

\bibitem{ICC}
P.~E. Shrout and J.~L. Fleiss, ``Intraclass correlations: uses in assessing
  rater reliability.,'' {\em Psychological bulletin}, vol.~86, no.~2, p.~420,
  1979.

\bibitem{li2018eac}
W.~Li, F.~Abtahi, Z.~Zhu, and L.~Yin, ``Eac-net: Deep nets with enhancing and
  cropping for facial action unit detection,'' {\em IEEE transactions on
  pattern analysis and machine intelligence}, vol.~40, no.~11, pp.~2583--2596,
  2018.

\bibitem{walecki2017deep}
R.~Walecki, V.~Pavlovic, B.~Schuller, M.~Pantic, {\em et~al.}, ``Deep
  structured learning for facial action unit intensity estimation,'' in {\em
  Proceedings of the IEEE Conference on Computer Vision and Pattern
  Recognition}, pp.~3405--3414, 2017.

\bibitem{SRERL}
G.~Li, X.~Zhu, Y.~Zeng, Q.~Wang, and L.~Lin, ``Semantic relationships guided
  representation learning for facial action unit recognition,'' in {\em
  Proceedings of the AAAI Conference on Artificial Intelligence}, vol.~33,
  pp.~8594--8601, 2019.

\bibitem{song2021uncertain}
T.~Song, L.~Chen, W.~Zheng, and Q.~Ji, ``Uncertain graph neural networks for
  facial action unit detection,'' in {\em Proceedings of the AAAI Conference on
  Artificial Intelligence}, vol.~35, pp.~5993--6001, 2021.

\bibitem{song2021hybrid}
T.~Song, Z.~Cui, W.~Zheng, and Q.~Ji, ``Hybrid message passing with
  performance-driven structures for facial action unit detection,'' in {\em
  Proceedings of the IEEE/CVF Conference on Computer Vision and Pattern
  Recognition}, pp.~6267--6276, 2021.

\bibitem{ME-GraphAU}
C.~Luo, S.~Song, W.~Xie, L.~Shen, and H.~Gunes, ``Learning multi-dimensional
  edge feature-based au relation graph for facial action unit recognition,''
  {\em arXiv preprint arXiv:2205.01782}, 2022.

\bibitem{yang2019facs3d}
L.~Yang, I.~O. Ertugrul, J.~F. Cohn, Z.~Hammal, D.~Jiang, and H.~Sahli,
  ``Facs3d-net: 3d convolution based spatiotemporal representation for action
  unit detection,'' in {\em 2019 8th International Conference on Affective
  Computing and Intelligent Interaction (ACII)}, pp.~538--544, IEEE, 2019.

\bibitem{AdamW}
I.~Loshchilov and F.~Hutter, ``Decoupled weight decay regularization,'' {\em
  arXiv preprint arXiv:1711.05101}, 2017.

\bibitem{Xavier}
X.~Glorot and Y.~Bengio, ``Understanding the difficulty of training deep
  feedforward neural networks,'' in {\em Proceedings of the thirteenth
  international conference on artificial intelligence and statistics},
  pp.~249--256, JMLR Workshop and Conference Proceedings, 2010.

\bibitem{large_batch}
P.~Goyal, P.~Doll{\'a}r, R.~Girshick, P.~Noordhuis, L.~Wesolowski, A.~Kyrola,
  A.~Tulloch, Y.~Jia, and K.~He, ``Accurate, large minibatch sgd: Training
  imagenet in 1 hour,'' {\em arXiv preprint arXiv:1706.02677}, 2017.

\bibitem{SGDR}
I.~Loshchilov and F.~Hutter, ``Sgdr: Stochastic gradient descent with warm
  restarts,'' {\em arXiv preprint arXiv:1608.03983}, 2016.

\bibitem{PyTorch}
A.~Paszke, S.~Gross, F.~Massa, A.~Lerer, J.~Bradbury, G.~Chanan, T.~Killeen,
  Z.~Lin, N.~Gimelshein, L.~Antiga, {\em et~al.}, ``Pytorch: An imperative
  style, high-performance deep learning library,'' {\em Advances in neural
  information processing systems}, vol.~32, 2019.

\bibitem{RandAug}
E.~D. Cubuk, B.~Zoph, J.~Shlens, and Q.~V. Le, ``Randaugment: Practical
  automated data augmentation with a reduced search space,'' in {\em
  Proceedings of the IEEE/CVF conference on computer vision and pattern
  recognition workshops}, pp.~702--703, 2020.

\bibitem{DropPath}
G.~Huang, Y.~Sun, Z.~Liu, D.~Sedra, and K.~Q. Weinberger, ``Deep networks with
  stochastic depth,'' in {\em European conference on computer vision},
  pp.~646--661, Springer, 2016.

\bibitem{LabelSmoothing}
C.~Szegedy, V.~Vanhoucke, S.~Ioffe, J.~Shlens, and Z.~Wojna, ``Rethinking the
  inception architecture for computer vision,'' in {\em Proceedings of the IEEE
  conference on computer vision and pattern recognition}, pp.~2818--2826, 2016.

\bibitem{mixup}
H.~Zhang, M.~Cisse, Y.~N. Dauphin, and D.~Lopez-Paz, ``mixup: Beyond empirical
  risk minimization,'' {\em arXiv preprint arXiv:1710.09412}, 2017.

\bibitem{cutmix}
S.~Yun, D.~Han, S.~J. Oh, S.~Chun, J.~Choe, and Y.~Yoo, ``Cutmix:
  Regularization strategy to train strong classifiers with localizable
  features,'' in {\em Proceedings of the IEEE/CVF international conference on
  computer vision}, pp.~6023--6032, 2019.

\bibitem{ISIR}
J.~Nicolle, K.~Bailly, and M.~Chetouani, ``Facial action unit intensity
  prediction via hard multi-task metric learning for kernel regression,'' in
  {\em 2015 11th IEEE International Conference and Workshops on Automatic Face
  and Gesture Recognition (FG)}, vol.~6, pp.~1--6, IEEE, 2015.

\bibitem{2DC}
D.~Linh~Tran, R.~Walecki, S.~Eleftheriadis, B.~Schuller, M.~Pantic, {\em
  et~al.}, ``Deepcoder: Semi-parametric variational autoencoders for automatic
  facial action coding,'' in {\em Proceedings of the IEEE International
  Conference on Computer Vision}, pp.~3190--3199, 2017.

\bibitem{G2RL}
Y.~Fan and Z.~Lin, ``G2rl: geometry-guided representation learning for facial
  action unit intensity estimation,'' in {\em Proceedings of the Twenty-Ninth
  International Conference on International Joint Conferences on Artificial
  Intelligence}, pp.~731--737, 2021.

\bibitem{RE-Net}
H.~Yang and L.~Yin, ``Re-net: A relation embedded deep model for au occurrence
  and intensity estimation,'' in {\em Proceedings of the Asian Conference on
  Computer Vision}, 2020.

\bibitem{FRL}
A.~Bulat, S.~Cheng, J.~Yang, A.~Garbett, E.~Sanchez, and G.~Tzimiropoulos,
  ``Pre-training strategies and datasets for facial representation learning,''
  {\em arXiv preprint arXiv:2103.16554}, 2021.

\bibitem{zhao2015loss}
H.~Zhao, O.~Gallo, I.~Frosio, and J.~Kautz, ``Loss functions for neural
  networks for image processing,'' {\em arXiv preprint arXiv:1511.08861}, 2015.

\end{thebibliography}

\end{document}